\documentclass[
  aps,
  pra,
  reprint,
  amsmath,
  amssymb,
  superscriptaddress,
  nofootinbib
]{revtex4-2}

\usepackage{graphicx}
\usepackage{dcolumn}
\usepackage{bm}
\usepackage{physics}
\usepackage{subcaption}
\usepackage{ragged2e}

\usepackage{booktabs}
\usepackage{multirow}
\usepackage{adjustbox}

\usepackage{hyperref}
\hypersetup{
    colorlinks=true,
    citecolor=blue,
    urlcolor=blue,
    linkcolor=blue
}

\begin{document}

\title{Neural Quantum Embedding for Multiclass Classification}

\author{Lu An}
\affiliation{School of Science, Zhejiang University of Science and Technology, Hangzhou 310023, China}
\author{Yuan-Hong Tao}
\thanks{Corresponding author: Yuan-Hong Tao, taoyuanhong12@126.com}
\affiliation{School of Science, Zhejiang University of Science and Technology, Hangzhou 310023, China}
\author{Peng-Fei Ma}
\affiliation{School of Science, Zhejiang University of Science and Technology, Hangzhou 310023, China}
\author{Shu-Hui Wu}
\affiliation{ZUST-CYU City of the Future Engineering Institute, Zhejiang University of Science and Technology, Hangzhou 310023, China}
\author{Shao-Ming Fei}
\affiliation{School of Mathematical Sciences, Capital Normal University, Beijing 100048, China}

\begin{abstract}
A multiclass neural quantum embedding (MNQE) method is proposed for multiclass classification, which enhances the separability of quantum states associated with different classes. We establish a rigorous connection between the trace distance between ensembles of quantum states from different classes and the lower bound of the empirical risk, providing insight into the role of quantum-state separability in classification performance. Experimental results on the MNIST and Fashion-MNIST datasets demonstrate that our MNQE significantly improves classification accuracy over existing quantum embedding methods in both noiseless and noisy environments. Furthermore, the resulting models exhibit improved generalization, expressivity and trainability.

\end{abstract}

\maketitle

\section{Introduction}

The development of noisy intermediate-scale quantum (NISQ) devices \cite{J. Preskill} has driven increasing interest in quantum machine learning as an application of quantum computing. By exploiting the representation capabilities of high-dimensional quantum systems and the intrinsic parallelism of quantum computation, quantum models have demonstrated potential for supervised learning and made substantial progress \cite{V, M. Schuld, C. Blank, H. Y. Huang, E. Peters}. Among various quantum models, quantum neural networks \cite{E. Farhi, I. Cong, A. Abbas, L. Liu, M. Boneberg} have been extensively investigated owing to their flexible parameterized architectures and favorable hardware compatibility. In the quantum neural networks driven by classical data, the quantum embedding \cite{Schuld} maps classical data into quantum states and constitutes the starting point of the quantum learning process. After quantum embedding, the original data are represented as quantum states in a Hilbert space, whose overall distribution directly affects the performance of the learning model, including its expressivity \cite{M. Schuld1}, generalization ability \cite{M. C. Caro1} and trainability \cite{S. Thanasilp}. 

In quantum classification, the separability of quantum states associated with different classes induced by the quantum embedding directly affects the classification performance. Such separability can be characterized by the trace distance \cite{C. W. Helstrom, J. Bae}. However, since completely positive trace-preserving maps cannot increase the trace distance between quantum states \cite{M. A. Nielsen}, the ability of quantum-circuit-based embeddings to enhance the separability of class-dependent quantum states is theoretically limited. Existing quantum embeddings predominantly employ fixed circuit architectures \cite{V, Schuld}, in which the mapping is predetermined and thus may not achieve optimal separability among quantum states from different classes. To address this issue, neural quantum embedding has been proposed as a data-driven approach for optimizing quantum embeddings \cite{T. Hur}. Its central idea is to introduce a classical neural network to perform a nonlinear transformation of the input data, thereby enhancing the separability between the embedded quantum states of two classes.

It has been demonstrated that neural quantum embedding can effectively improve the performance of quantum classification models \cite{T. Hur, H. F. Liu, Y. Kim}. However, the existing studies of neural quantum embedding have focused primarily on binary classification, with the optimization objective formulated in terms of the discrimination between quantum states associated with two classes. In multiclass tasks, the quantum states corresponding to multiple classes exhibit more complex discrimination relationships after quantum embedding. How to characterize the separability among multiclass quantum states and its relationship to classification performance, and how to further design quantum embedding optimization methods suitable for multiclass tasks, remain insufficiently explored.

Motivated by the above considerations, we propose a neural quantum embedding framework for multiclass classification. We first establish the relationship between the separability of class-dependent quantum states and the empirical risk in a $K$-class quantum classification task from the perspective of quantum-state discrimination, and derive a lower bound on the empirical risk. This result provides a theoretical characterization of the intrinsic relationship between quantum-state separability and classification performance. Based on the theoretical analysis of the lower bound on the empirical risk, we design a multiclass quantum embedding optimization strategy. By constructing a fidelity-based quantum embedding loss function and using it to train a classical neural network to learn the feature representation prior to quantum embedding, the proposed strategy jointly optimizes both intraclass quantum-state concentration and interclass quantum-state separability, thereby enhancing the separability of quantum states associated with different classes.

We perform systematic experiments on the MNIST \cite{Y. Lecun} and Fashion-MNIST \cite{H. Xiao} multiclass datasets using the PennyLane platform, and evaluate the effect of multiclass neural quantum embedding on quantum classification models under both ideal and noisy quantum conditions. The experimental results show that, compared with models without multiclass neural quantum embedding, the proposed method effectively improves the classification performance across four four-class tasks, achieving an average increase of 61.49 percentage points in test accuracy. Compared with existing quantum embedding optimization methods, the proposed multiclass neural quantum embedding achieves an average improvement of 12.29 percentage points in test accuracy. Beyond classification accuracy, we further investigate the effects on quantum classification models from the perspectives of model expressivity, local effective dimension \cite{A. Abbas, O. Berezniuk, A. Abbas1} and trainability. The results collectively demonstrate that, compared with existing quantum embedding methods, multiclass neural quantum embedding can effectively improve classification performance and trainability while exhibiting strong robustness under noisy conditions.

The paper is organized as follows. Section~\ref{sec:2} introduces quantum neural networks and the relevant preliminaries. Section~\ref{sec:3} derives a lower bound on the empirical risk for $K$-class quantum classification and presents the proposed multiclass neural quantum embedding method, including the design of the loss function and quantum embedding strategy. Section~\ref{sec:4} validates the effectiveness of the proposed method through numerical simulations. Section~\ref{sec:5} further investigates the effects of multiclass neural quantum embedding on the generalization ability, expressivity and trainability of quantum classification models. Section~\ref{sec:6} concludes the paper.

\section{Preliminaries}
\label{sec:2}

This section introduces the basic theory of quantum supervised learning, with particular emphasis on quantum neural network models and their classification processes, providing the theoretical foundation for the subsequent development of a multiclass neural quantum embedding method.

Quantum neural networks (QNN) constitute a typical class of quantum supervised learning models. Supervised learning uses labeled training samples to learn a mapping between the input and output spaces. Given a training set $S=\{(x_i,y_i)\}_{i=1}^{N}$, the learning process can be formulated as finding a prediction function $f$ in the function space $\mathcal{F}$ that can effectively predict the outputs of unseen samples. The performance of the prediction function is typically quantified by the risk function,
\begin{equation}
R(f)=\mathbb{E}\bigl[l(f(X),Y)\bigr],
\end{equation}
where $l(\cdot)$ denotes the loss function, and $(X,Y)$ follows an unknown data distribution. Since the true risk is generally intractable to evaluate directly, the empirical risk is typically used as the optimization objective in practical training:
\begin{equation}
R_N(f)=
\frac{1}{N}
\sum_{i=1}^{N}
l(f(x_i),y_i).
\end{equation}
According to the empirical risk minimization principle, the prediction function is obtained by solving
\begin{equation}
\hat f
=
\arg\min_{f\in\mathcal{F}}
R_N(f).
\end{equation}
Quantum supervised learning algorithms aim to exploit the computational capabilities of quantum devices to find prediction functions with improved performance.

The computational process of a QNN consists of quantum embedding, parameterized unitary evolution and measurement-based discrimination, as described below.

(1) Quantum embedding: Given a classical input $x \in \mathbb{R}^m$, a quantum embedding map $\Phi(x)$ encodes the input into a quantum state
$|x\rangle = \Phi(x)\,|0\rangle^{\otimes n}$,
thereby embedding the classical data into a Hilbert space.

(2) Parameterized unitary evolution: A parameterized unitary operator $U(\theta)$ is applied to the embedded state, yielding
$|x,\theta\rangle = U(\theta)|x\rangle$,
where $\theta$ denotes the trainable parameters.

(3) Measurement-based discrimination: The evolved quantum state is measured using an observable $O$, and the model output is given by
\begin{equation}
f(x;\theta)=\langle x|U^\dagger(\theta),O,U(\theta)|x\rangle.
\end{equation}
The parameterized unitary evolution and measurement can be combined into an effective measurement operator
$E(\theta)=U^\dagger(\theta)OU(\theta)$. The model output can then be equivalently written as $f(x;\theta)=\langle x|E(\theta)|x\rangle$.
The resulting expectation value is used in the learning task. The model parameters are optimized using gradient descent or its variants to minimize the empirical risk.

A QNN determines the class of an input sample based on the measurement outcomes in classification tasks. From the perspective of quantum measurement theory, this process can be regarded as a quantum state discrimination problem. For a $K$-class classification task, the set $\{E_i(\theta)\}_{i=1}^{K}$ induced by the parameters $\theta$ constitutes a positive-operator-valued measure (POVM), satisfying $E_i(\theta)\geq 0$ and $\sum_{i=1}^{K}E_i(\theta)=I$ with $I$ being the identity operator. Given an input sample $x$, the conditional probability that the model assigns the sample to class $i$ is
\begin{equation}
P(i|x,\theta)=\langle x|E_i(\theta)|x\rangle,
\end{equation}
where $E_i(\theta)$ denotes the measurement operator associated with the class $i$. This expression gives the probability of obtaining the outcome corresponding to class $i$ when the quantum state $|x\rangle$ is measured with $E_i(\theta)$. The predicted class is determined according to the maximum-probability rule:
\begin{equation}
\hat{y}=\arg\max_i P(i|x,\theta).
\end{equation}

The neural quantum embedding method proposed in Ref. \cite{T. Hur} introduces a classical neural network prior to quantum embedding to apply a nonlinear transformation to the input data, thereby increasing the distance between the embedded quantum states of the two classes and improving the performance of the quantum classification model. However, this method is formulated for binary classification and does not directly account for the relationships among quantum states associated with multiple classes. To address multiclass quantum classification tasks, we propose a multiclass neural quantum embedding method that optimizes the representation of quantum states across multiple classes.

\section{MULTICLASS NEURAL QUANTUM EMBEDDING (MNQE)}
\label{sec:3}

We first derive a lower bound on the empirical risk for $K$-class quantum classification tasks, and then design the overall framework of multiclass neural quantum embedding based on the bound.

\subsection{Design Rationale for Multiclass Neural Quantum Embedding}
\label{sec:3A}

The classification performance of a model for classification task is quantified by the empirical risk. 
Within the QNN framework introduced in Section~\ref{sec:2}, a natural loss function can be defined as the misclassification probability:
\begin{equation}
l(f(x;\theta),y)
=
1-P(y|x,\theta),
\end{equation}
where $P(y|x,\theta)$ denotes the probability that the sample $x$ is correctly assigned to its true class $y$. Given a training set $S=\{(x_i,y_i)\}_{i=1}^{N}$, the empirical risk can be written as
\begin{equation}
L_s=\frac{1}{N}\sum_{i=1}^{N}\bigl[1-P(y_i|x_i,\theta)\bigr].
\end{equation}
Suppose that the training set contains $K$ classes, with $N_i$ samples in class $i$. The quantum-state ensemble corresponding to class $i$ can be represented by the statistical mixture
$\rho_i=\frac{1}{N_i}\sum_{y_j=i}|x_j\rangle\langle x_j|$,
with prior probability $p_i=N_i/N$. For the $K$-class quantum classification task, we derive the following lower bound on the empirical risk, see the derivation in Appendix~\ref{app:risk_lower_bound}, 
\begin{equation}
L_s
\ge
1-\frac{1+2\sum_{i<j}D_{\mathrm{tr}}(p_i\rho_i,p_j\rho_j)}{K}.
\label{eq:risk_lower_bound}
\end{equation}
where $D_{\mathrm{tr}}(\cdot,\cdot)$ denotes the trace distance \cite{C. W. Helstrom, J. Bae}. For $K=2$, this expression naturally reduces to the empirical-risk lower bound for binary classification \cite{T. Hur}.

It follows from Eq.~\eqref{eq:risk_lower_bound} that the lower bound on the empirical risk in quantum classification is determined by the trace distances between the quantum-state ensembles of different classes. Larger interclass trace distances imply greater distinguishability between quantum states from different classes and a lower empirical-risk lower bound. Notably, this lower bound is determined entirely by the initial quantum embedding circuit and is independent of the specific structure of the subsequent parameterized unitary transformation $U(\theta)$. Moreover, as the trace distance satisfies the contractivity property: for any positive trace-preserving map $\Lambda$ \cite{Siudzinska},
\begin{equation}
D_{\mathrm{tr}}\!\left(\Lambda(\rho_i),\Lambda(\rho_j)\right)
\le
D_{\mathrm{tr}}(\rho_i,\rho_j),
\end{equation}
the subsequent quantum operations cannot increase the trace distance between quantum-state ensembles of different classes. The performance of quantum classification is largely constrained by the class structure established by the initial quantum embedding.

Since the lower bound on the empirical risk is controlled by the interclass trace distances, a key objective of quantum embedding is to maximize the distances between quantum states associated with different classes, thereby reducing the theoretical lower bound on the empirical risk. In the noisy intermediate-scale quantum regime, noise and other nonunitary processes further contract the trace distance \cite{M. A. Nielsen}, reducing the separation between classes. It is therefore desirable to construct a trainable and data-dependent quantum embedding that enhances interclass distinguishability and thereby improves the risk lower bound. Several studies have introduced parameterized quantum gates into conventional quantum embedding circuits to construct trainable unitary embedding schemes \cite{S. Lloyd, T. Hubregtsen, J. R Glick, L. Xu}. Although such approaches provide greater flexibility in quantum embeddings, their optimization capability remains limited by several factors. First, introducing parameterized structures substantially increases the circuit depth and gate count, which not only increases the computational complexity but also makes the system more susceptible to noise \cite{S. Wang}. Second, these models are prone to barren plateaus during training \cite{J. R. McClean, M. Cerezo2}, which may severely limit optimization efficiency and scalability. More importantly, previous studies have shown that the ability of trainable unitary embeddings to increase the maximum trace distance between embedded quantum states is fundamentally limited \cite{T. Hur}. Consequently, the existing quantum embedding methods generally cannot guarantee a sufficiently large separation between data ensembles of different classes in Hilbert space, thereby limiting the achievable performance of quantum models in classification tasks.

\subsection{Multiclass Neural Quantum Embedding Procedure}

\begin{figure*}[!t]
    \centering
    \begin{subfigure}{0.58\textwidth}
        \centering
        \includegraphics[width=\textwidth]{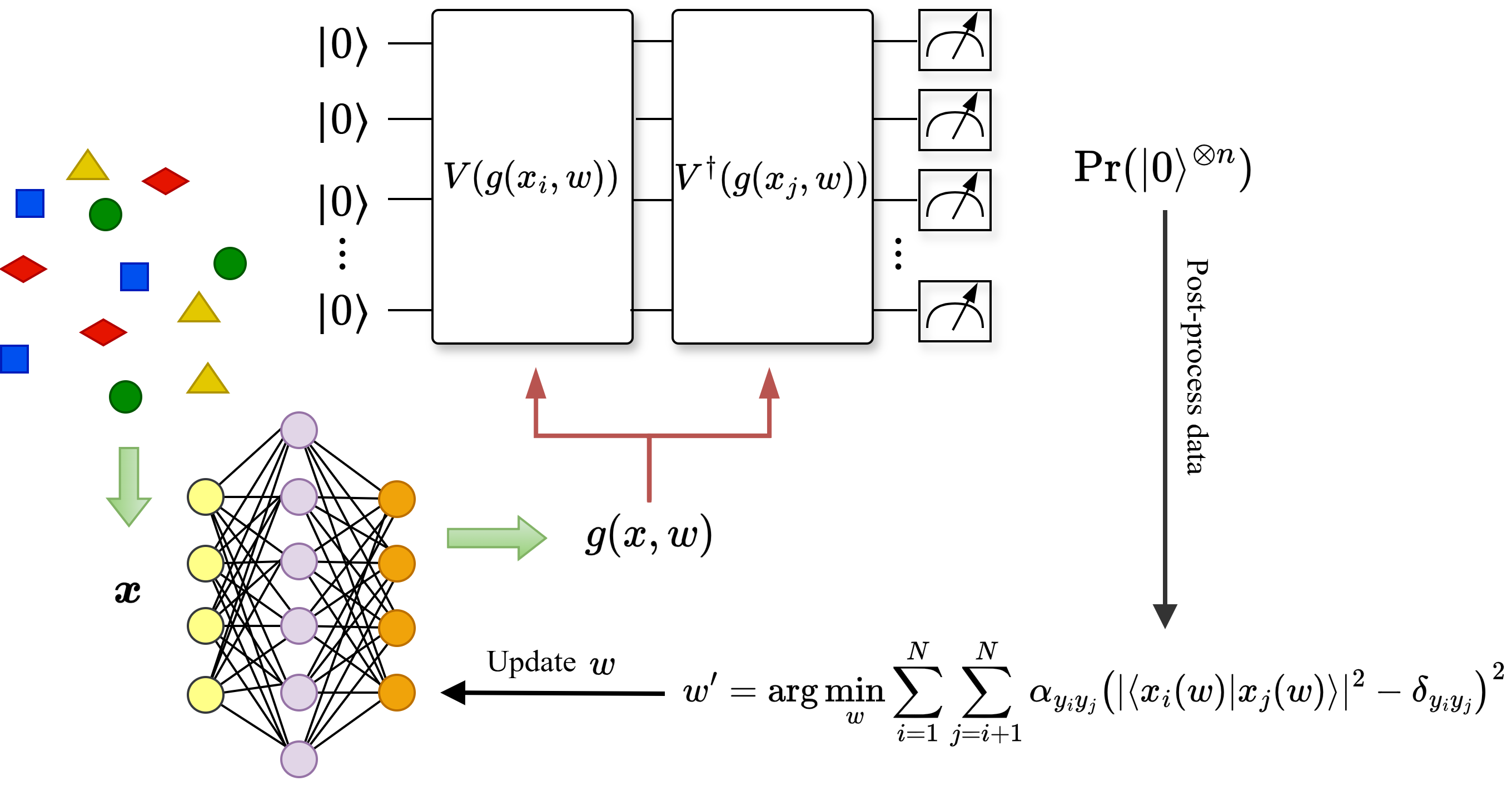}
        \caption{}
        \label{fig:workflow_mnqe}
    \end{subfigure}
    \hfill
    \begin{subfigure}{0.40\textwidth}
        \centering
        \includegraphics[width=\textwidth]{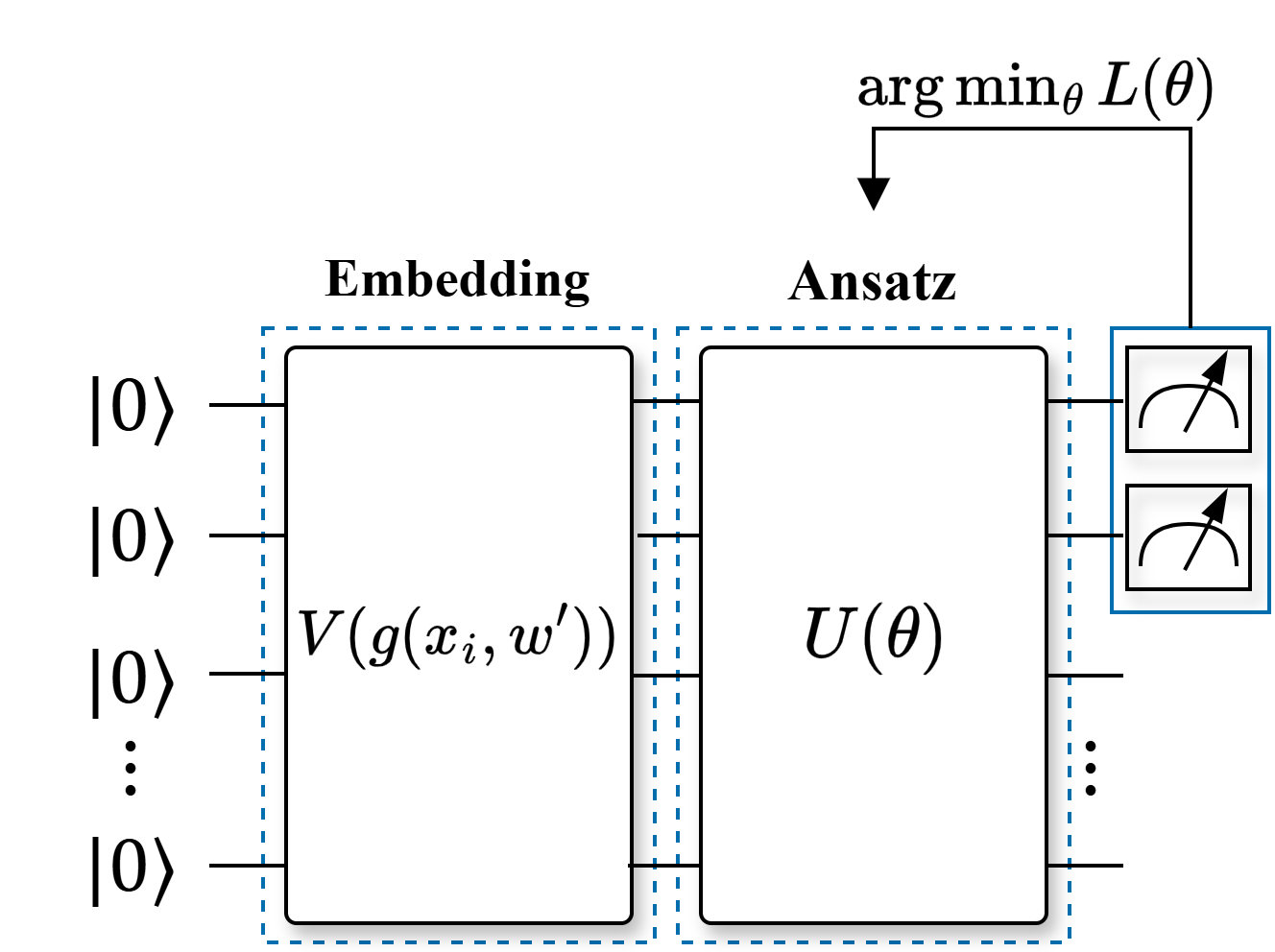}
        \caption{}
        \label{fig:workflow_qnn}
    \end{subfigure}
    \caption{\justifying Workflow of the MNQE-based quantum neural network classification algorithm. (a) MNQE training stage: a classical neural network is optimized by minimizing the loss function $L$ based on quantum-state fidelity to obtain the optimal classical parameters $w^{\prime}$, such that the trace distances between the quantum-state ensembles of different classes are maximized after quantum embedding by $V(g(x,w^{\prime}))$. (b) Quantum neural network classification training stage: after data embedding, the quantum parameters $\theta$ are optimized to adjust the measurement operators by minimizing the classification loss function $L(\theta)$.}
    \label{fig:Workflow}
\end{figure*}

Based on the theoretical analysis on the empirical risk lower bound in Section~\ref{sec:3A}, we introduce the multiclass neural quantum embedding (MNQE) method. A classical neural network is introduced before the quantum embedding to optimize the data representation, thereby overcoming the limitation on interclass distinguishability imposed by completely positive trace-preserving (CPTP) maps in conventional quantum embedding. The overall procedure is shown in Fig.~\ref{fig:workflow_mnqe}.

For an input sample $x\in\mathbb{R}^{m}$, MNQE first applies a classical neural network to perform a feature transformation
$g:\mathbb{R}^{m}\times\mathbb{R}^{r}\rightarrow\mathbb{R}^{m'}$.
Denote $w\in\mathbb{R}^{r}$ the trainable parameters. The transformed features $g(x,w)$ are then fed into the quantum embedding circuit $V$, yielding the quantum state,
\begin{equation}
|x\rangle = V(g(x,w))|0\rangle^{\otimes n}.
\end{equation}
Thus, the overall mapping can be written as $x \rightarrow g(x,w) \rightarrow |x\rangle$. In addition to increasing the interclass trace distance, MNQE can perform effective feature compression during quantum embedding by choosing $m' < m$, thereby avoiding the use of classical dimensionality-reduction methods, such as PCA or autoencoders, before embedding and accommodating the limited number of controllable qubits available in current quantum devices.

For the multiclass setting, the ideal objective of MNQE is to maximize the distinguishability between quantum-state ensembles of different classes, as quantified by the trace distance. However, the trace distance is computationally expensive to evaluate and is therefore difficult to use directly for training. We thus introduce an implicit fidelity-based loss function,
\begin{equation}
L =
\sum_{i=1}^{N}
\sum_{j=i+1}^{N}
\alpha_{y_i y_j}
\left(
|\langle x_i | x_j \rangle|^2
-
\delta_{y_i y_j}
\right)^2,
\label{eq:loss}
\end{equation}
where $|\langle x_i | x_j \rangle|^2$ denotes the fidelity between the $i$th and $j$th quantum-state samples, and $\delta_{y_i y_j}$ is the Kronecker delta, which equals 1 for $y_i=y_j$ and 0 otherwise. The weight coefficient $\alpha_{y_i y_j}$ controls the contributions of intraclass and interclass sample pairs,
\begin{equation}
\alpha_{y_i y_j} =
\begin{cases}
\alpha_{ss}, & y_i = y_j \\
\alpha_{sd}, & y_i \neq y_j
\end{cases}
\end{equation}
where $\alpha_{ss}$ controls intraclass compactness and $\alpha_{sd}$ controls interclass separation. This loss function jointly optimizes the similarity of intraclass quantum states and the distinguishability of interclass quantum states according to the class labels. During training, the fidelities can be efficiently evaluated using the SWAP test \cite{C. J. Garcia-Escartin, L. Cincio} or obtained directly by measuring the overlap between quantum states, as illustrated by the circuit in Fig.~\ref{fig:workflow_mnqe}. Moreover, the quantum-state fidelity is closely related to the trace distance. Minimizing the loss function $L$ is equivalent to enhancing the distinguishability between the samples from different classes, see the detailed relationship and derivation in Appendix~\ref{app:loss_trace_distance}.

\begin{figure*}[!t]
    \centering
    \begin{subfigure}{0.49\textwidth}
        \centering
        \includegraphics[width=\textwidth]{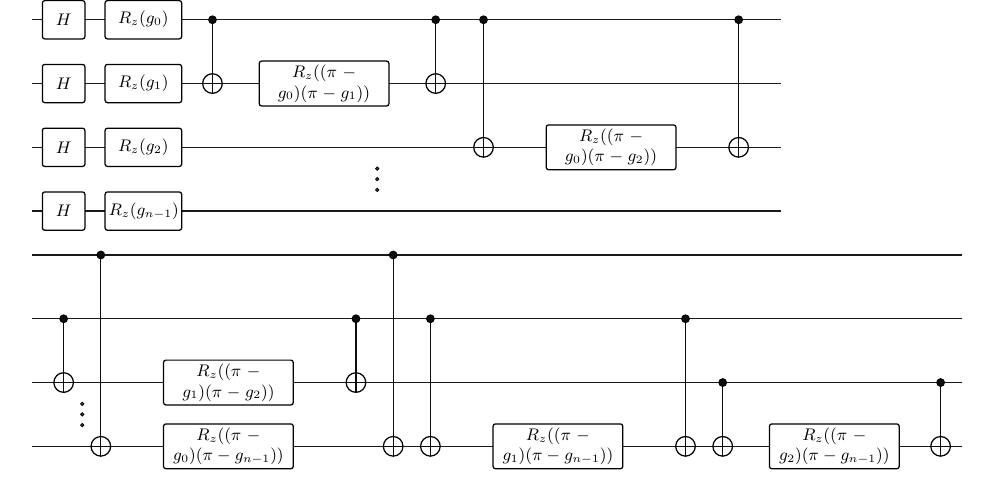}
        \caption{}
        \label{fig:circuit1}
    \end{subfigure}
    \hfill
    \begin{subfigure}{0.49\textwidth}
        \centering
        \includegraphics[width=\textwidth]{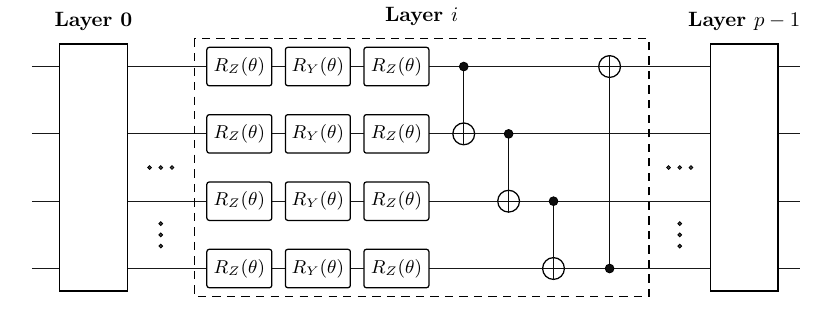}
        \caption{}
        \label{fig:circuit2}
    \end{subfigure}
    \caption{\justifying (a) Schematic of the quantum circuit for the single-layer trainable ZZ feature map $V(g(x_i,w))$. For clarity, the complete quantum circuit is divided into two parts and shown separately above and below. The feature map encodes the input vector $x_i$ using the mapping functions $g_j \in g(x_i,w)$. (b) Schematic of the structure of $U(\theta)$ in Fig.~\ref{fig:workflow_qnn}. The superscripts and subscripts of the parameters are omitted for simplicity.}
    \label{fig:circuit}
\end{figure*}

To construct the quantum embedding layer of MNQE, we employ the representative ZZ feature map \cite{V} as the underlying quantum embedding structure, as shown in Fig.~\ref{fig:circuit1}. The corresponding unitary transformation is defined as
\begin{equation}
V(\phi(x))
=
\left[
\exp\left(
i\sum_i \phi_i(x)Z_i
+
i\sum_{i,j}\phi_{ij}(x)Z_iZ_j
\right)
H^{\otimes n}
\right]^L,
\end{equation}
where $H$ denotes the Hadamard gate, $Z_i$ denotes the Pauli-$Z$ operator acting on the $i$th qubit, and $L$ is the number of embedding layers. For $L>1$, the corresponding quantum kernel is considered to exhibit classical intractability \cite{V} and is therefore widely used as a benchmark embedding structure in quantum machine learning. In the conventional ZZ feature map, the mapping functions are typically chosen as $\phi_i(x)=x_i$ and $\phi_{ij}(x)=\frac{(\pi-x_i)(\pi-x_j)}{2}$ \cite{V, A. Abbas}. However, such fixed mapping functions lack a clear design rationale and may not adapt well to the data distribution of a specific task. MNQE instead replaces the fixed mapping functions with a trainable classical neural network, which learns data-dependent feature transformations to adaptively optimize the quantum embedding.

In our proposed MNQE method, the structure of the quantum embedding circuit remains fixed, with no additional trainable quantum parameters. During training, the quantum processor evaluates the fidelities between embedded quantum states, while the classical computer updates the neural-network parameters, enabling classical-quantum co-optimization. After training, the embedding layer is used in the training of the quantum neural network ansatz layer, as illustrated in Fig.~\ref{fig:workflow_qnn}.

\section{MULTICLASS EXPERIMENTAL VALIDATION OF MNQE}
\label{sec:4}

To evaluate the effectiveness and performance of MNQE, we compare the effects of different quantum embedding methods on classification performance across several four-class classification tasks, including the baseline quantum embedding, MNQE, and existing trainable quantum embedding methods.

\subsection{Experimental Design}

The experiments are implemented on the PennyLane platform. The quantum classification model adopts a quantum neural network architecture consisting of a quantum embedding layer and a quantum classification layer, where the latter is implemented using an 8-qubit hardware-efficient variational quantum circuit. The MNIST \cite{Y. Lecun} and Fashion-MNIST \cite{H. Xiao} datasets are used for the classification tasks. For MNIST, the digit classes $\{0,1,2,3\}$ and $\{0,1,6,8\}$ are considered, while for Fashion-MNIST, the classes $\{0,1,2,3\}$ and $\{0,1,5,9\}$ are considered. In total, four four-class classification tasks are constructed to evaluate the feature learning capability and classification performance of the proposed method.

The hardware-efficient variational quantum circuit consists of rotation and entangling modules. The rotation module uses $R_Z$ and $R_Y$ gates to construct an arbitrary single-qubit Euler rotation:
\begin{equation}
U_{q,i}
=
R_Z\left(\theta_{q,i}^{2}\right)
R_Y\left(\theta_{q,i}^{1}\right)
R_Z\left(\theta_{q,i}^{0}\right),
\label{eq:ansatz}
\end{equation}
where $q$ and $i$ denote the qubit and circuit-layer indices, respectively, and $\theta_{q,i}^{0}$, $\theta_{q,i}^{1}$, and $\theta_{q,i}^{2}$ are trainable parameters. The entangling module consists of CNOT gates arranged in a ring topology. The rotation and entangling modules are alternately stacked for $p$ layers to form the hardware-efficient variational quantum circuit used in the experiments \cite{L. Liu}, as shown in Fig.~\ref{fig:circuit2}.

The experiments are conducted on noiseless and noisy quantum simulators, and consist of three main stages: (1) the embedding layers are trained using MNQE and trainable quantum feature mapping (TQFM) \cite{L. Xu}, respectively; (2) the classification layers of the quantum neural network (QNN) are trained using the trained embedding layers and the embedding layer without MNQE, i.e., the conventional ZZ feature map, respectively; and (3) the model performance is evaluated and compared using accuracy, precision, recall and F1 score.

\subsection{Comparative Analysis of MNQE Classification Performance}

We compare the proposed MNQE method with the conventional ZZ feature map and the trainable quantum feature mapping (TQFM) proposed in Ref. \cite{L. Xu}. Although Ref. \cite{L. Xu} primarily investigates trainable quantum kernels for quantum classifiers and discusses the potential application of the method to quantum neural networks (QNN), we introduce TQFM into the QNN framework as a quantum embedding method for comparison based on its quantum feature map structure. The ZZ feature map uses the fixed analytical mapping functions $\phi_i(x)=x_i$ and $\phi_{ij}(x)=\frac{(\pi-x_i)(\pi-x_j)}{2}$ to encode classical data into quantum states.

Because the number of qubits that can be reliably controlled by current quantum devices is limited, high-dimensional data generally require dimensionality reduction before being fed into a quantum embedding circuit. To address this issue, we consider two implementations of MNQE. In PCA-MNQE, principal component analysis (PCA) \cite{I. T. Jolliffe} is first applied to reduce the dimensionality of the input data, after which the resulting low-dimensional features are fed into a classical neural network for feature learning. In MNQE, the input and output dimensions of the neural network are adjusted to enable end-to-end feature compression and learning. Both methods produce 8-dimensional feature vectors, which are used as the rotation-gate parameters of the 8-qubit ZZ feature map circuit. The main difference between the two approaches lies in their input processing: PCA-MNQE constructs a low-dimensional representation using PCA, whereas MNQE directly learns the feature representation from the original $28\times28$ images without additional classical dimensionality reduction. To ensure a fair comparison, all methods use the same circuit configuration, except for the embedding circuit of TQFM, which is adjusted according to its original design. The architecture of the classical neural network and the implementation details of the quantum circuits used in MNQE are provided in Appendix~\ref{app:Details}.

The TQFM method used for comparison performs feature learning by optimizing a parameterized quantum circuit to generate quantum states. The quantum states are constructed using the following layered parameterized quantum circuit:
\begin{equation}
|\tilde{x}\rangle=
\prod_{l=1}^{L}
\left[
V(\phi(x))
\exp\left(
i\sum_i\theta_i^lY_i
+i\sum_{i,j}\theta_{i,j}^lY_iY_j
\right)
\right]
|0\rangle^{\otimes n},
\end{equation}
where $V(\phi(x))$ denotes the ZZ feature map in Eq.~\eqref{eq:ansatz}, $i$ and $j$ denote the qubit indices, and $\theta_i^l$ and $\theta_{i,j}^l$ are trainable parameters corresponding to the single-qubit rotation and two-qubit entangling terms, respectively. Based on the parameterized quantum states described above, TQFM employs the loss function defined in Ref. \cite{L. Xu}:
\begin{equation}
E(\theta)
=
1-\frac{1}{K}
\sum_{k=1}^{K}
\frac{1}{N_k}
\sum_{s=1}^{N_k}
\left|\langle y_k|\tilde{x}_s^k\rangle\right|^2 ,
\end{equation}
where $K$ denotes the number of classes, $N_k$ denotes the number of samples in class $k$, $|y_k\rangle$ denotes the computational-basis label state corresponding to class $k$, and $|\tilde{x}_s^k\rangle$ denotes the quantum state generated by the parameterized quantum circuit for the $s$th sample in class $k$.

\begin{figure*}[!t]
    \centering
    \begin{subfigure}[t]{0.49\textwidth}
        \centering
        \includegraphics[width=\textwidth]{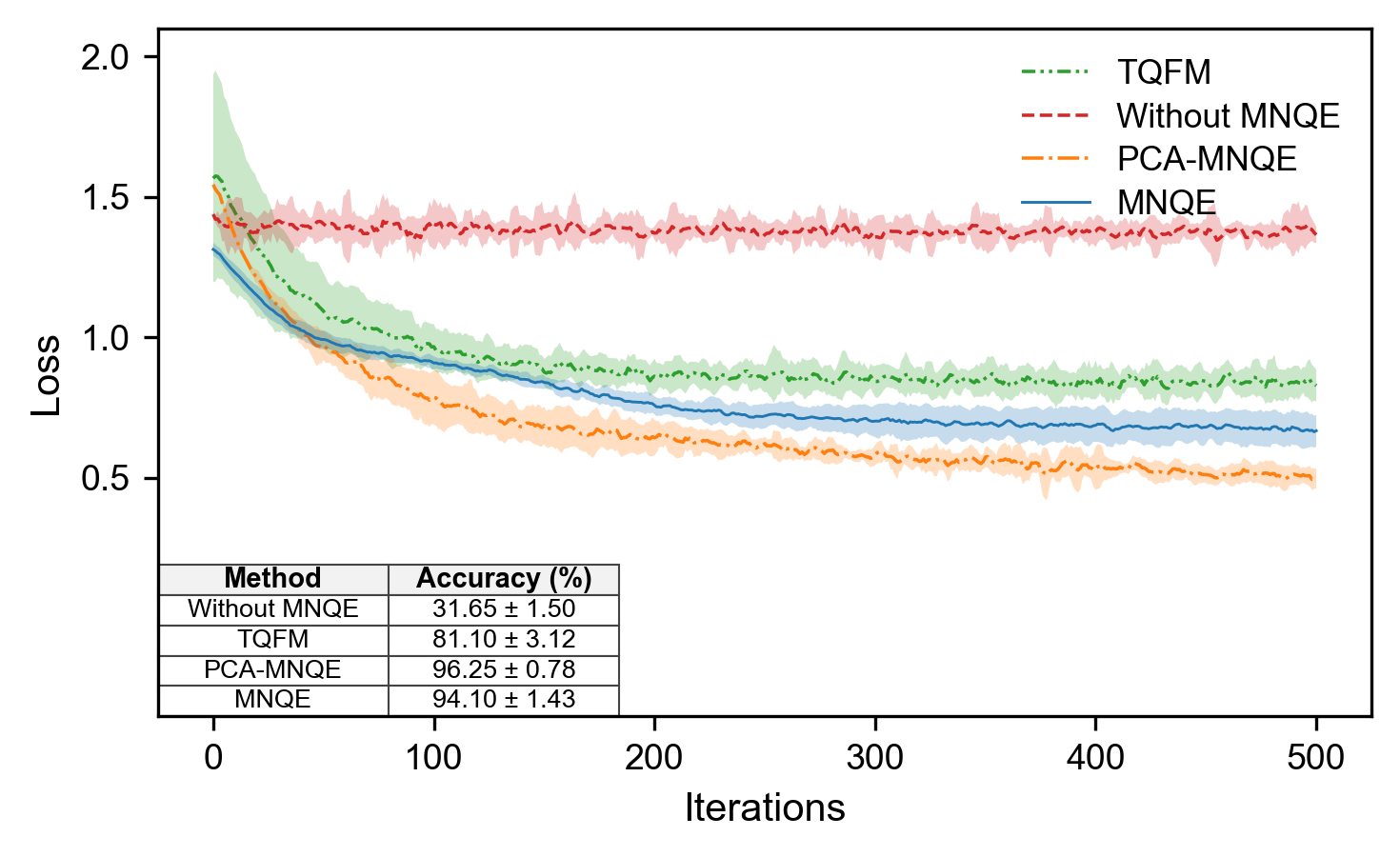}
        \caption{}
        \label{fig:2a}
    \end{subfigure}
    \hfill
    \begin{subfigure}[t]{0.49\textwidth}
        \centering
        \includegraphics[width=\textwidth]{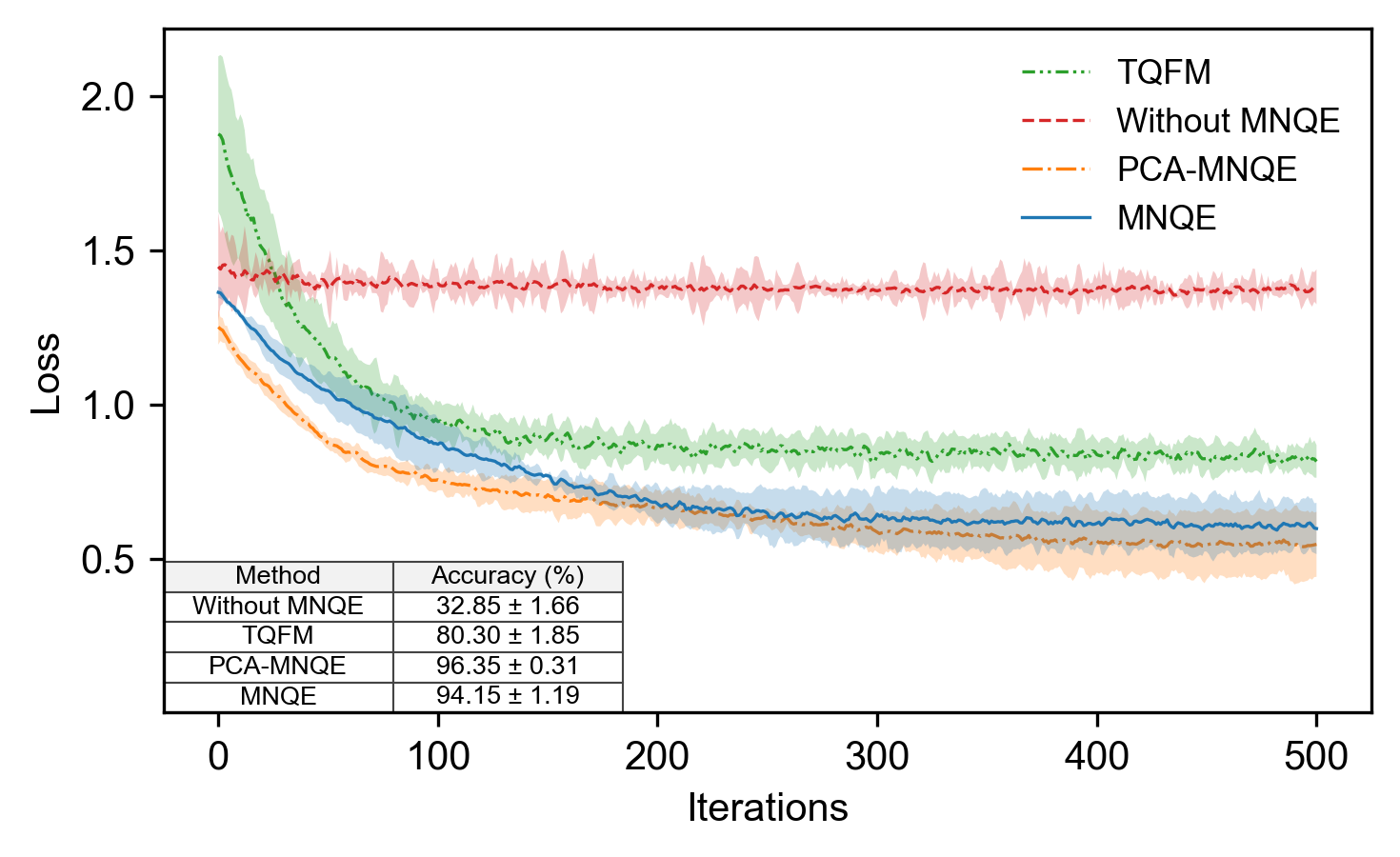}
        \caption{}
        \label{fig:2b}
    \end{subfigure}

    \vspace{0.3cm}

    \begin{subfigure}[t]{0.49\textwidth}
        \centering
        \includegraphics[width=\textwidth]{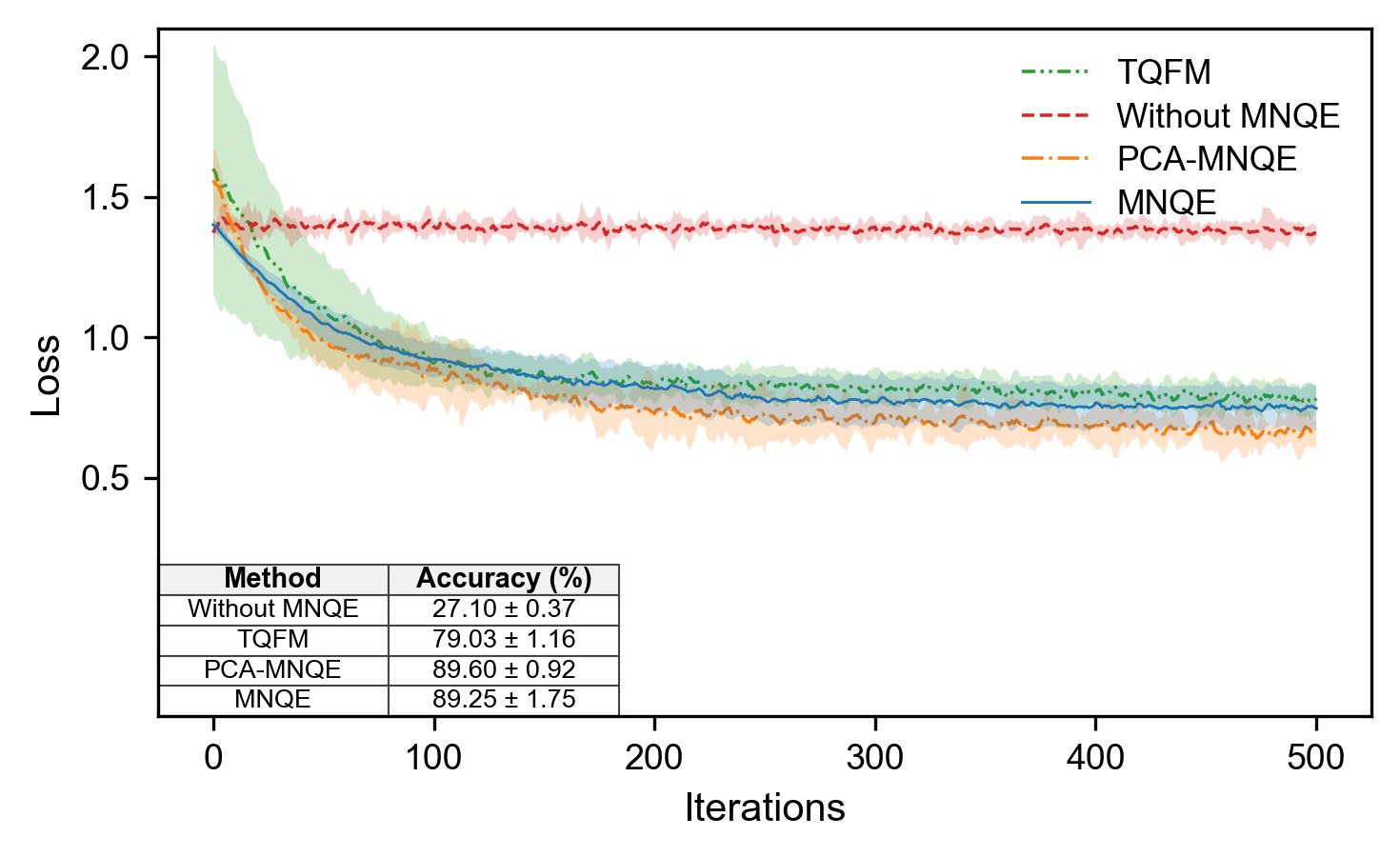}
        \caption{}
        \label{fig:2c}
    \end{subfigure}
    \hfill
    \begin{subfigure}[t]{0.49\textwidth}
        \centering
        \includegraphics[width=\textwidth]{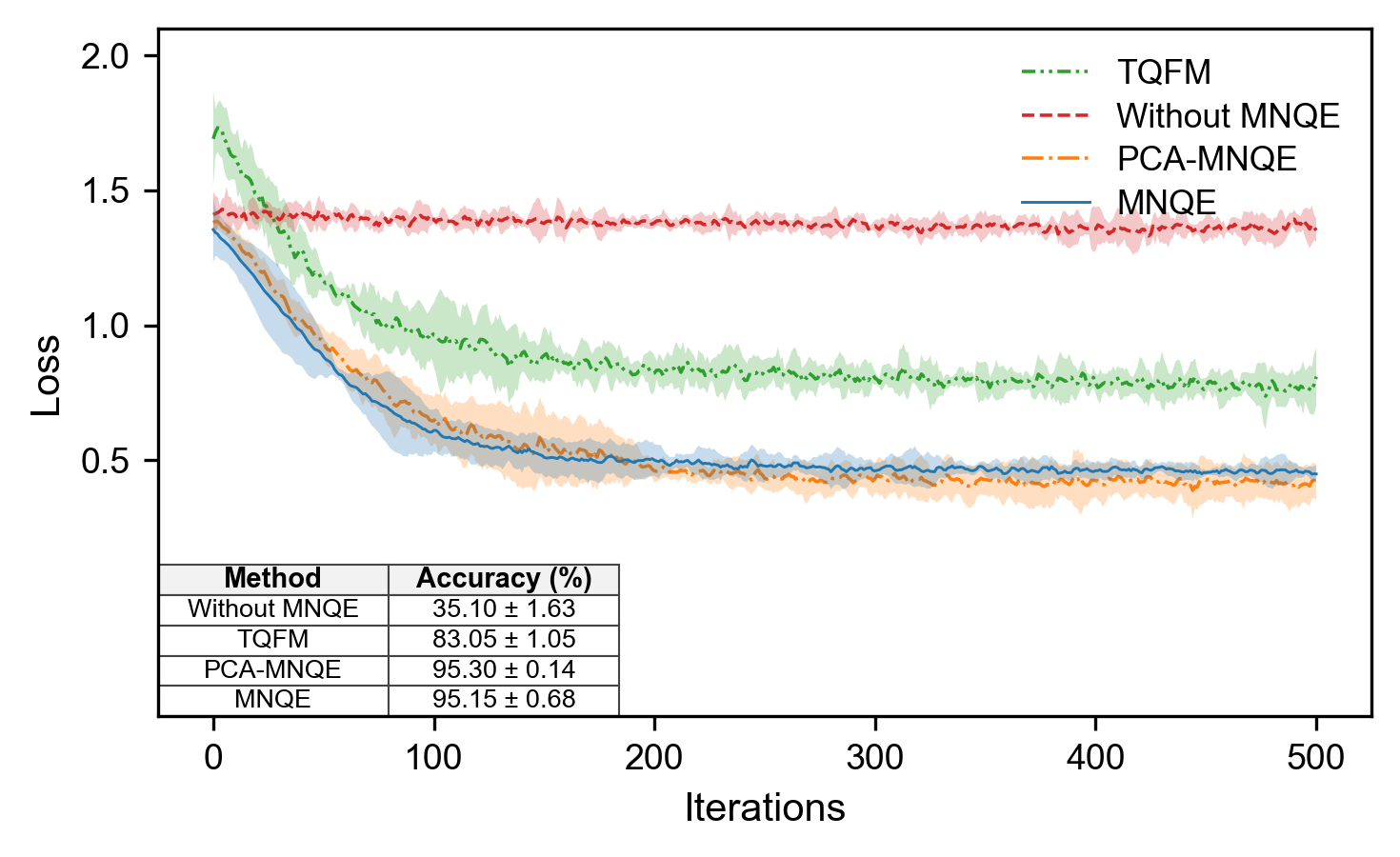}
        \caption{}
        \label{fig:2d}
    \end{subfigure}
    \caption{\justifying Training-loss curves of quantum neural networks for multiclass classification tasks under noiseless conditions. The cross-entropy loss is used as the optimization objective during training. The red dashed, blue solid, orange dash-dotted, and green dash-double-dotted curves represent the average training losses of the models using Without MNQE, MNQE, PCA-MNQE, and TQFM, respectively. Panels (a) and (b) correspond to the $\{0123\}$ and $\{0168\}$ class combinations of the MNIST dataset, respectively, while panels (c) and (d) correspond to the $\{0123\}$ and $\{0159\}$ class combinations of the Fashion-MNIST dataset, respectively. Each curve is obtained by averaging the results of five independent training runs with random parameter initialization; the shaded regions indicate one standard deviation above and below the mean. The tables show the mean and standard deviation of the test-set classification accuracy for each model.}
    \label{fig:combined}
\end{figure*}

Fig.~\ref{fig:combined} shows the simulation results of the quantum neural networks under noiseless conditions. All models use the cross-entropy loss as the training objective during the classification stage. The red dashed, blue solid, orange dash-dotted, and green dash-double-dotted curves represent the average training-loss curves of the quantum neural networks using Without MNQE, MNQE, PCA-MNQE, and TQFM, respectively. The average loss is calculated from five independent training runs with random parameter initialization for each method, and the shaded regions indicate one standard deviation from the mean. The test-set accuracies of the different methods are listed in the table in Fig.~\ref{fig:combined}. As shown in the figure, the training loss of Without MNQE remains relatively high throughout the training process and shows no clear decreasing trend, indicating that the fixed ZZ feature map provides limited feature representations for the classification tasks considered here. In contrast, the training losses of MNQE, PCA-MNQE, and TQFM gradually decrease with the number of iterations and eventually converge. Compared with Without MNQE, the other three methods achieve lower final training losses and higher classification accuracies. In particular, the two models using MNQE-trained features achieve better performance on the test sets, with test accuracies exceeding 90\%, outperforming TQFM and Without MNQE. Across the four four-class classification tasks, MNQE improves the average test accuracy by 61.49 and 12.29 percentage points compared with Without MNQE and TQFM, respectively. This improvement can be attributed to the ability of MNQE to learn task-specific feature representations in a data-driven manner, resulting in better separability of samples from different classes in the quantum-state space. This reduces the optimization difficulty of the subsequent classifier, ultimately leading to improved classification performance.

\subsection{Analysis of Trace Distance in MNQE}

\begin{figure*}[!t]
\centering

\begin{subfigure}[b]{0.28\textwidth}
    \centering
    \includegraphics[width=\textwidth]{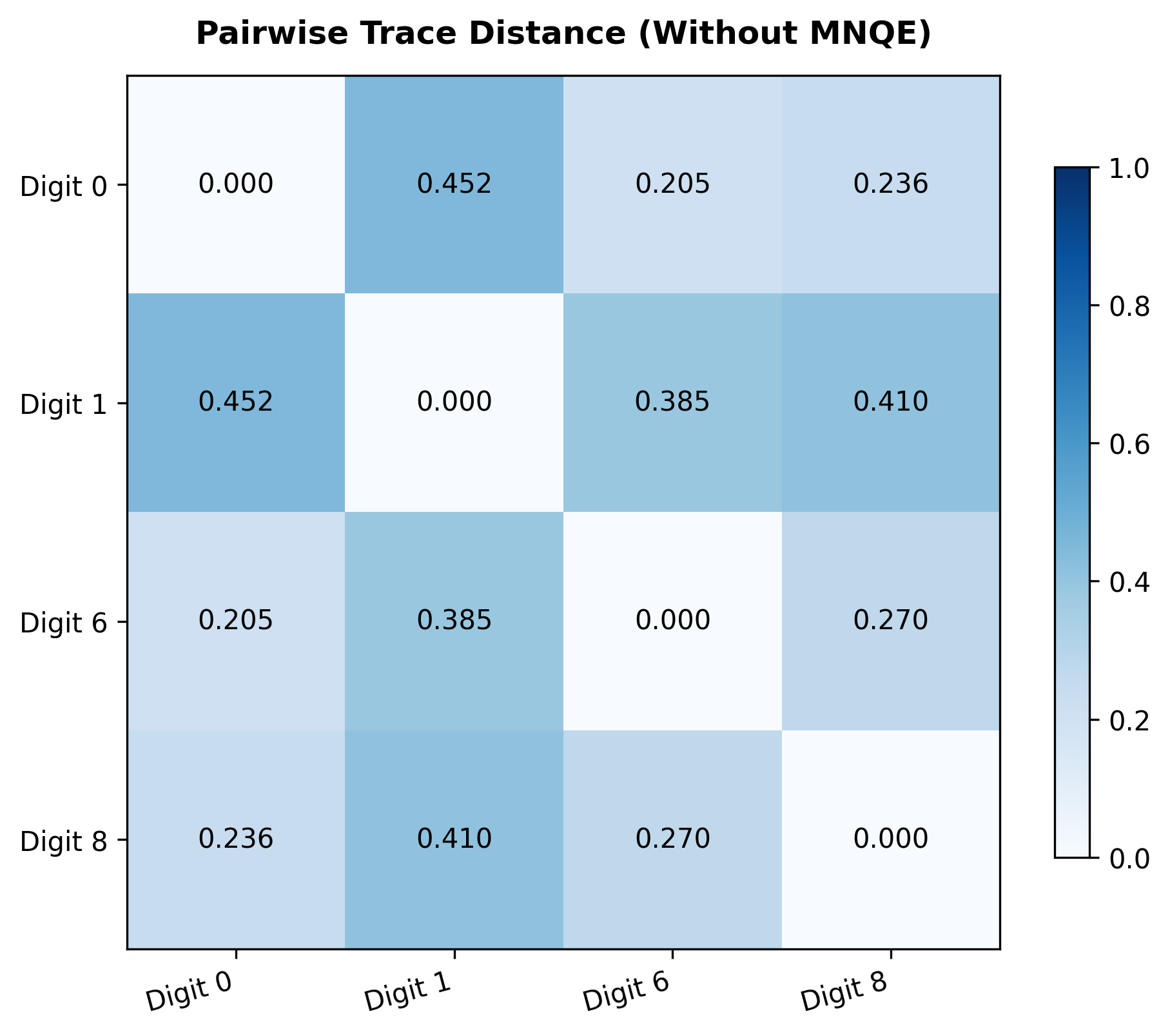}
    \caption{}
    \label{fig:a}
\end{subfigure}
\hfill
\begin{subfigure}[b]{0.28\textwidth}
    \centering
    \includegraphics[width=\textwidth]{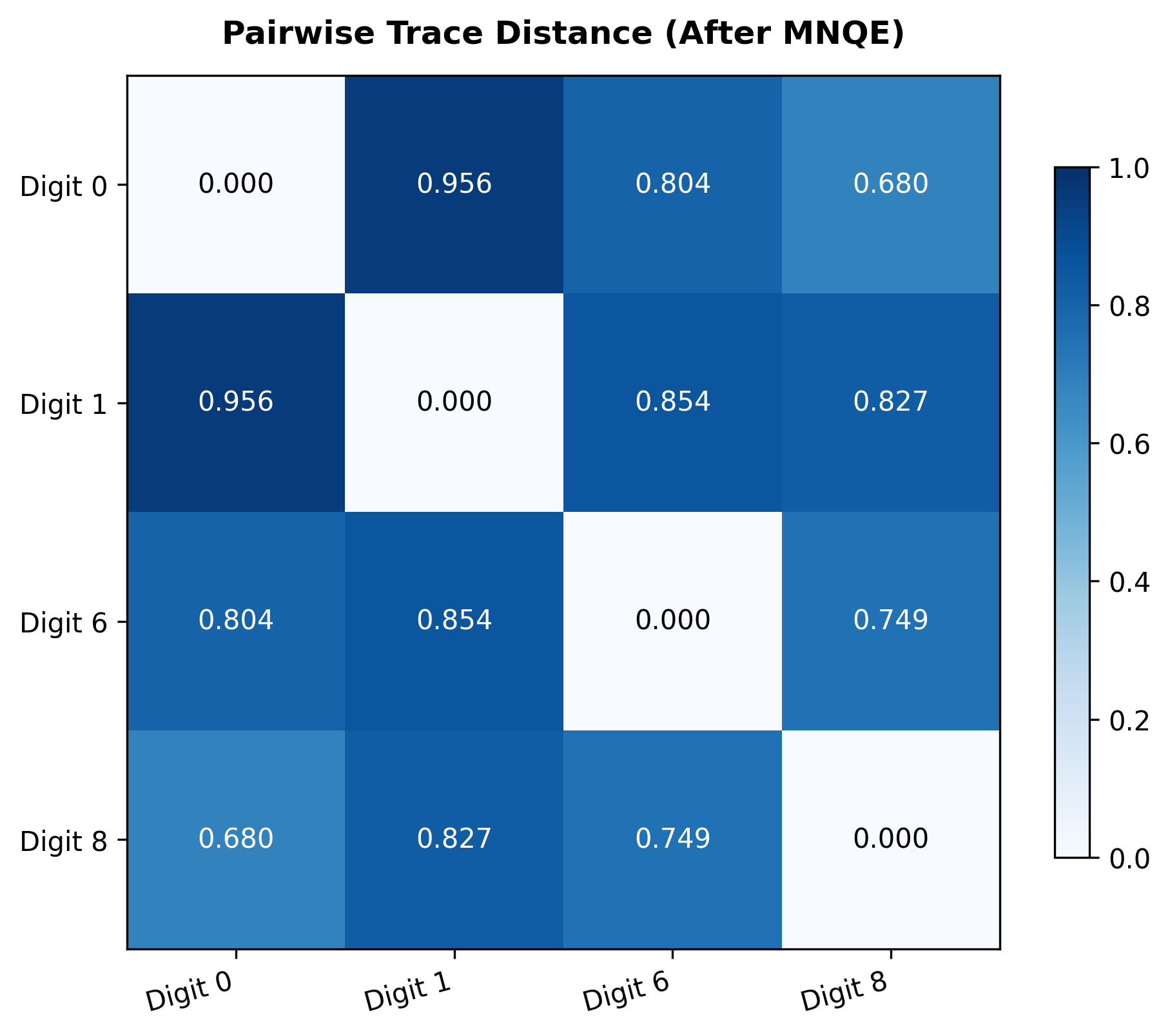}
    \caption{}
    \label{fig:b}
\end{subfigure}
\hfill
\begin{subfigure}[b]{0.37\textwidth}
    \centering
    \includegraphics[width=\textwidth]{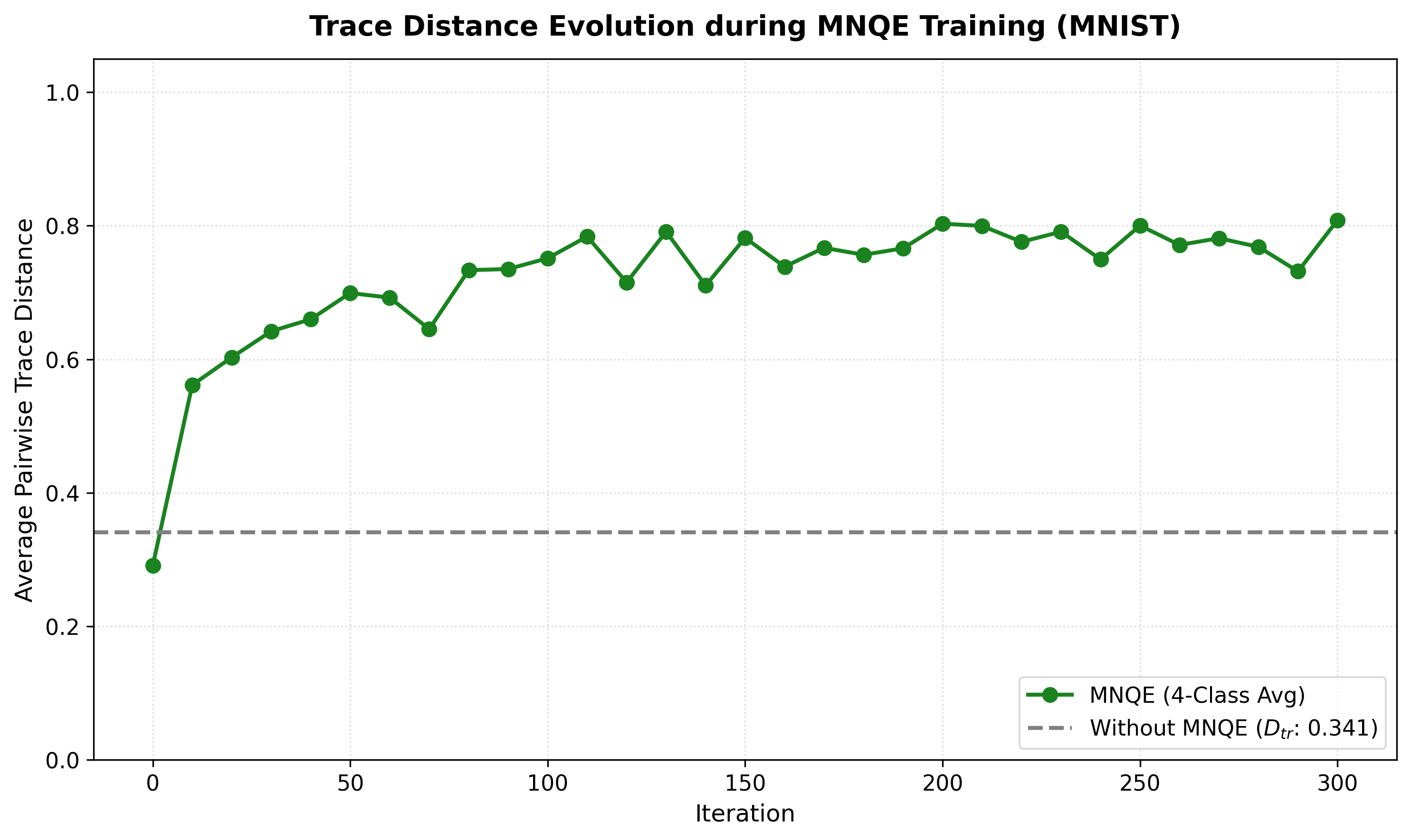}
    \caption{}
    \label{fig:c}
\end{subfigure}

\vspace{0.3cm}

\begin{subfigure}[b]{0.28\textwidth}
    \centering
    \includegraphics[width=\textwidth]{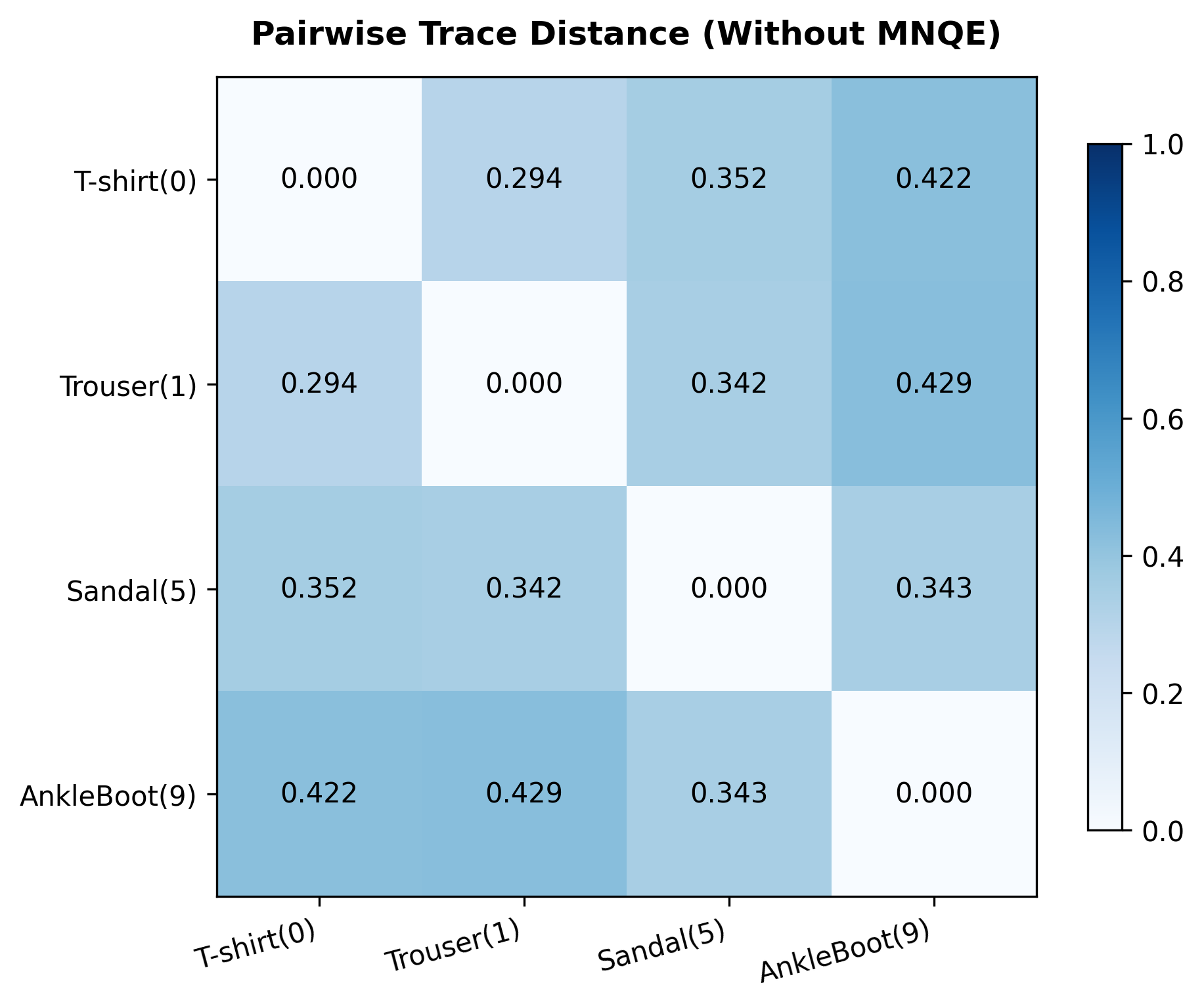}
    \caption{}
    \label{fig:d}
\end{subfigure}
\hfill
\begin{subfigure}[b]{0.28\textwidth}
    \centering
    \includegraphics[width=\textwidth]{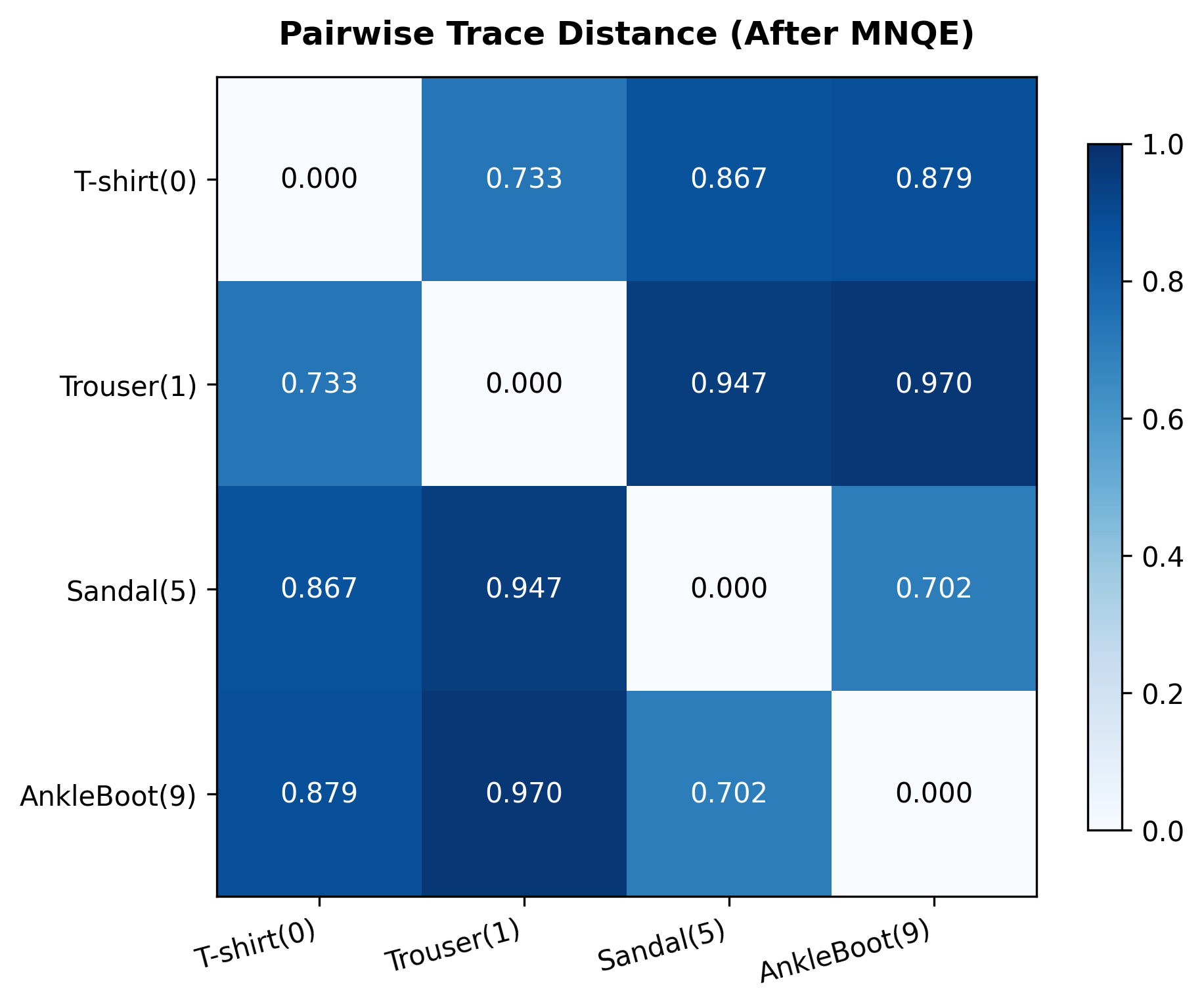}
    \caption{}
    \label{fig:e}
\end{subfigure}
\hfill
\begin{subfigure}[b]{0.37\textwidth}
    \centering
    \includegraphics[width=\textwidth]{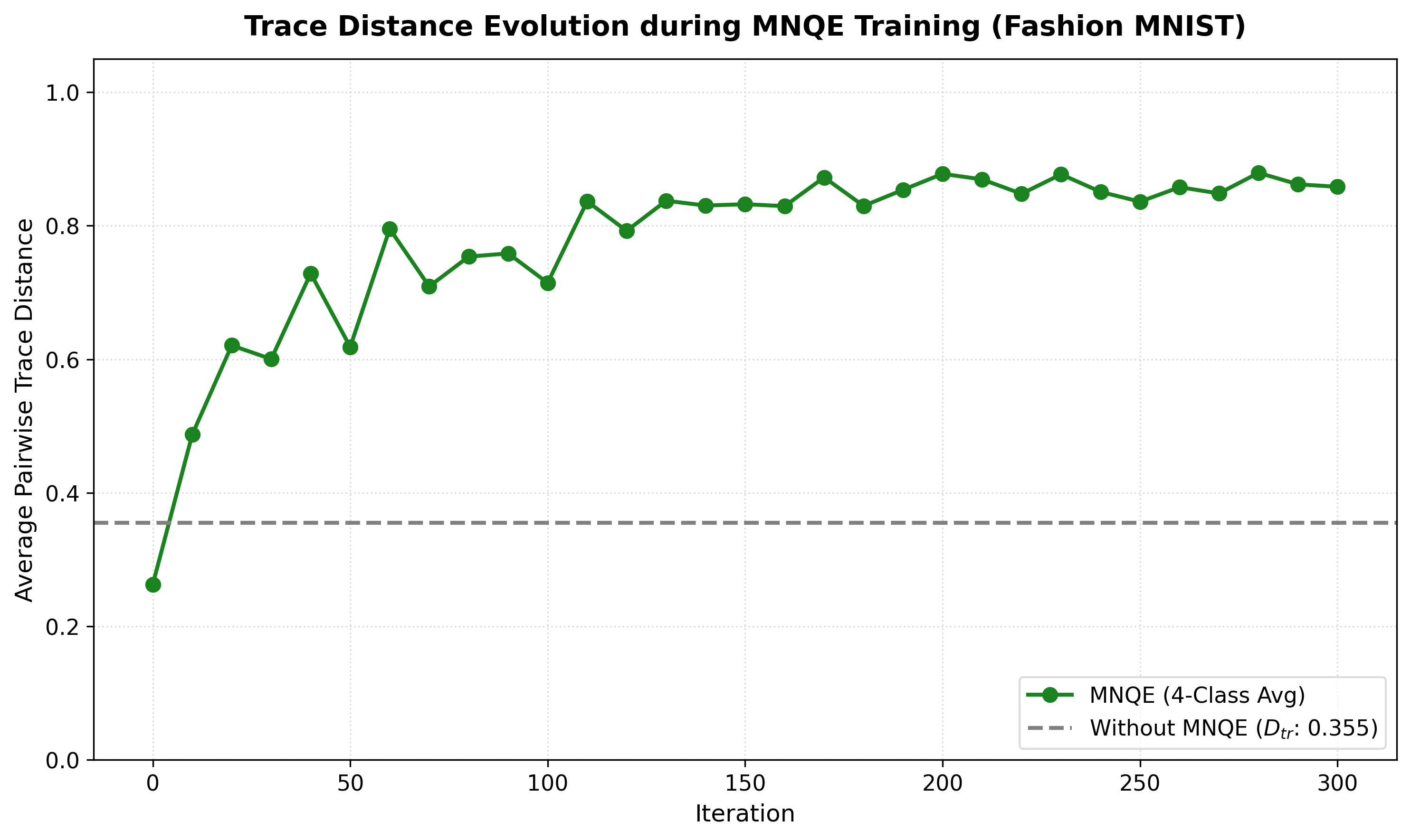}
    \caption{}
    \label{fig:f}
\end{subfigure}

\caption{\justifying Effect of MNQE on the trace distances between quantum-state ensembles of different classes. Panels (a) and (b) show the heat maps of the interclass trace distance matrices for the MNIST $\{0168\}$ dataset without and with MNQE, respectively, while panel (c) shows the evolution of the average trace distance with the number of iterations during MNQE training on MNIST $\{0168\}$. Panels (d) and (e) show the corresponding interclass trace distance matrices for the Fashion-MNIST $\{0159\}$ dataset without and with MNQE, respectively, while panel (f) shows the evolution of the average trace distance during MNQE training on Fashion-MNIST $\{0159\}$. The average trace distance is defined as the mean of the trace distances over all class pairs.}
\label{fig:distance}
\end{figure*}

We experimentally compare the trace distances between quantum states from different classes after quantum embedding with and without MNQE. First, the average density matrix is constructed for each class from its samples, and the trace distance between the average density matrices of any two classes is then calculated as \( D_{\mathrm{tr}}(\rho_i,\rho_j) = \frac{1}{2}\left\| \rho_i-\rho_j \right\|_1 \), where $\rho_i$ and $\rho_j$ denote the average density matrices corresponding to classes $i$ and $j$, respectively. The average of the trace distances over all class pairs is defined as the average trace distance $\overline{D}_{\mathrm{tr}}$:
\begin{equation}
\overline{D}_{\mathrm{tr}}
= \frac{2}{K(K-1)}
\sum_{i<j} D_{\mathrm{tr}}(\rho_i,\rho_j).
\end{equation}
The average trace distance characterizes the overall change in the distances between quantum states from different classes during MNQE training. Fig.~\ref{fig:distance} shows the heat maps of the pairwise trace distances $D_{\mathrm{tr}}$ before and after MNQE training for the MNIST $\{0168\}$ and Fashion-MNIST $\{0159\}$ datasets, together with the corresponding evolution of $\overline{D}_{\mathrm{tr}}$ during training. After embedding the original data using the feature map, the average trace distances between different classes are 0.341 and 0.355 for the two datasets, respectively. After MNQE training, these values increase to 0.812 and 0.858, respectively, indicating a substantial enhancement in the separability of the embedded quantum states across different classes. Correspondingly, the empirical risk lower bounds decrease from 0.494 and 0.484 to 0.141 and 0.107, respectively.

\subsection{Robustness Analysis of MNQE}

\begin{figure*}[!t]
    \centering
    \begin{subfigure}{0.49\textwidth}
        \centering
        \includegraphics[width=\textwidth]{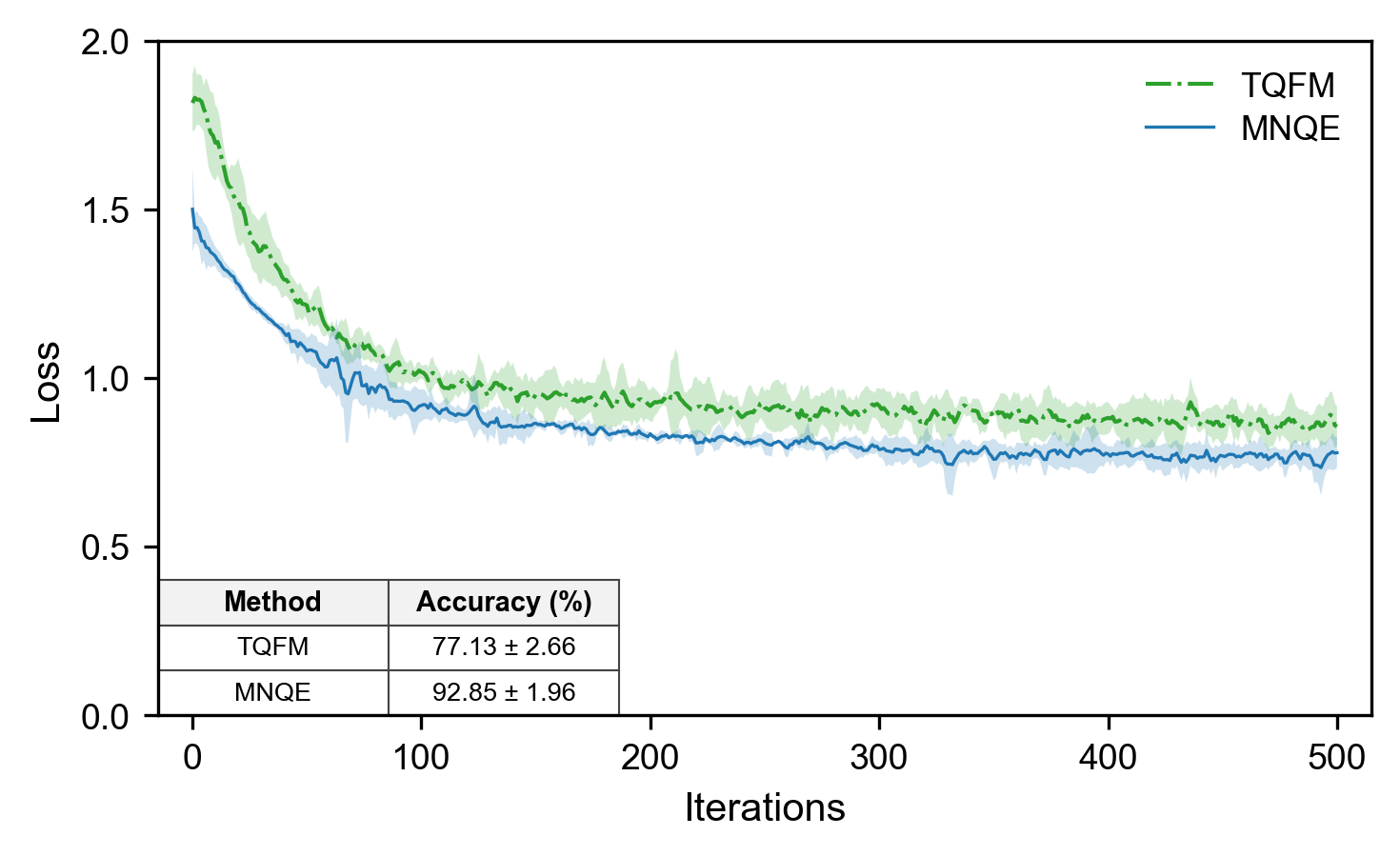}
        \caption{}
        \label{fig:noise1}
    \end{subfigure}
    \hfill
    \begin{subfigure}{0.49\textwidth}
        \centering
        \includegraphics[width=\textwidth]{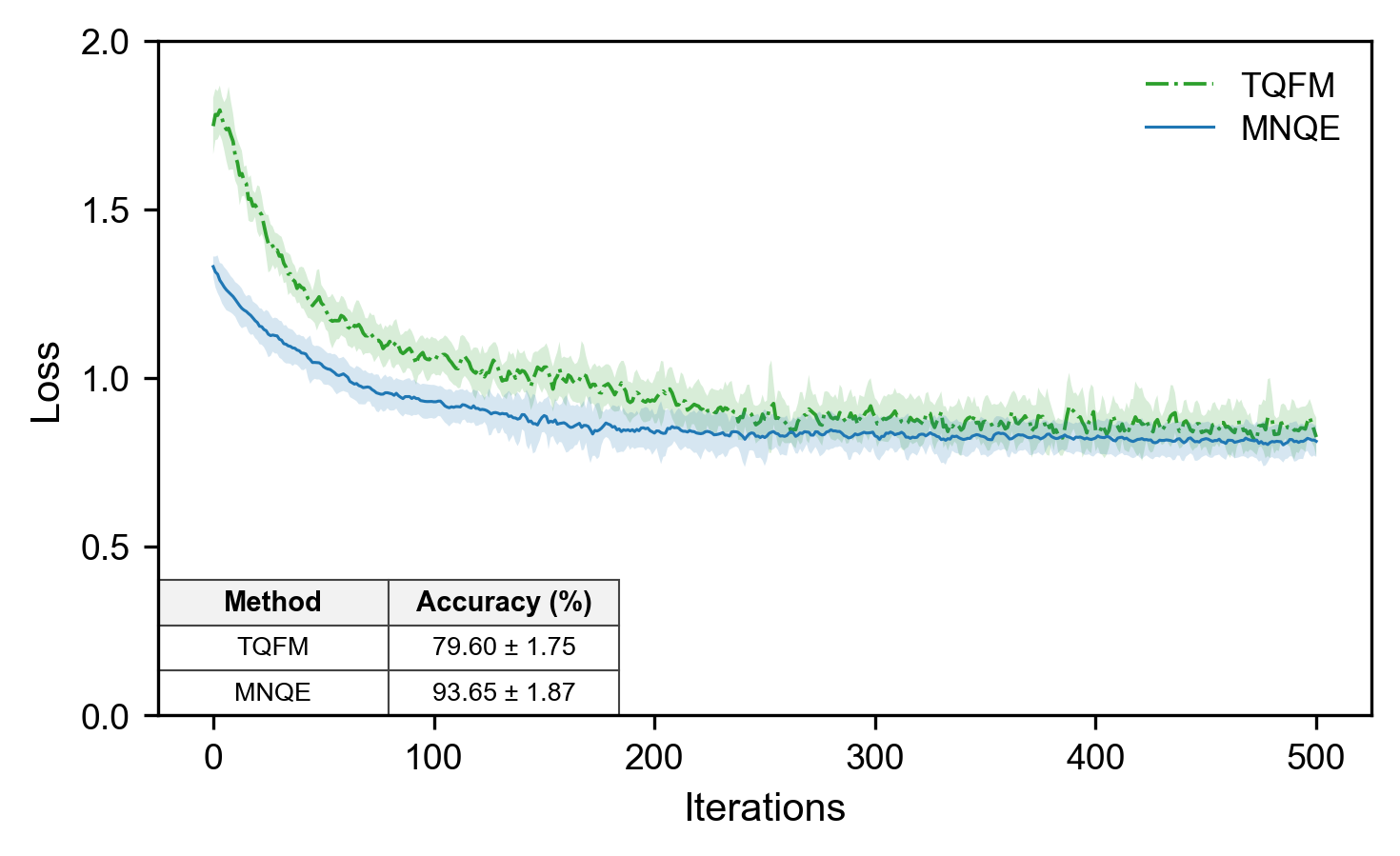}
        \caption{}
        \label{fig:noise2}
    \end{subfigure}
    \caption{\justifying Comparison of the robustness of MNQE and TQFM under Gaussian noise perturbations. Gaussian noise with a standard deviation of 0.05 is introduced into the input feature space normalized to $[0,1]$. Panels (a) and (b) correspond to the MNIST $\{0123\}$ and $\{0168\}$ datasets, respectively. The blue solid and green dash-dotted curves represent MNQE and TQFM, respectively. The curves show the mean training loss obtained from three independent training runs with random parameter initialization, and the shaded regions indicate one standard deviation. The tables report the mean and standard deviation of the test-set classification accuracy for each model.}
    \label{fig:noise}
\end{figure*}

To evaluate the robustness of the proposed method in the presence of noise, we introduce Gaussian noise with a standard deviation of 0.05 into the input feature space normalized to $[0,1]$ for the MNIST $\{0123\}$ and $\{0168\}$ classification tasks. The models using TQFM and MNQE are then evaluated and compared. In these experiments, both the embedding and Ansatz layers use three repeated circuit layers. Fig.~\ref{fig:noise} shows the mean training-loss curves obtained from three independent runs under noisy conditions, with the shaded regions indicating one standard deviation from the mean. The results show that MNQE maintains better classification performance in the presence of noise. In terms of classification accuracy, MNQE achieves 92.85\% and 93.65\% on the $\{0123\}$ and $\{0168\}$ tasks, respectively, whereas TQFM achieves only 77.13\% and 79.60\%. This performance advantage is attributed primarily to the larger initial trace distances and lower effective circuit depth, which improve the stability of the model under noisy conditions. Moreover, compared with the noiseless setting, MNQE exhibits a relatively small decrease in classification accuracy, indicating its robustness to noise perturbations.

\begin{figure*}[!t]
    \centering
    \begin{subfigure}[b]{0.57\textwidth}
        \centering
        \includegraphics[width=\textwidth]{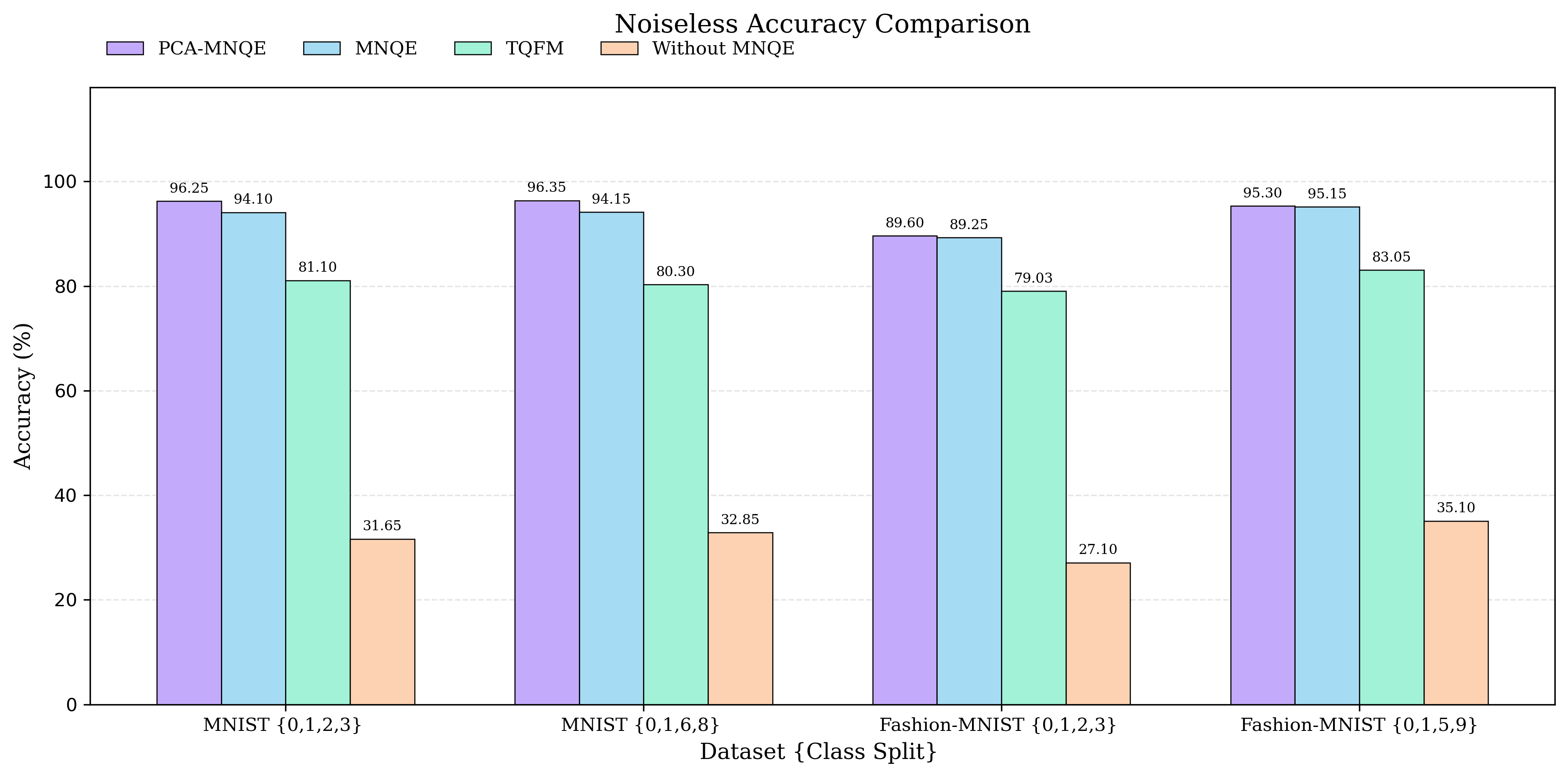}
        \caption{}
        \label{fig:15a}
    \end{subfigure}
    \hfill
    \begin{subfigure}[b]{0.38\textwidth}
        \centering
        \includegraphics[width=\textwidth]{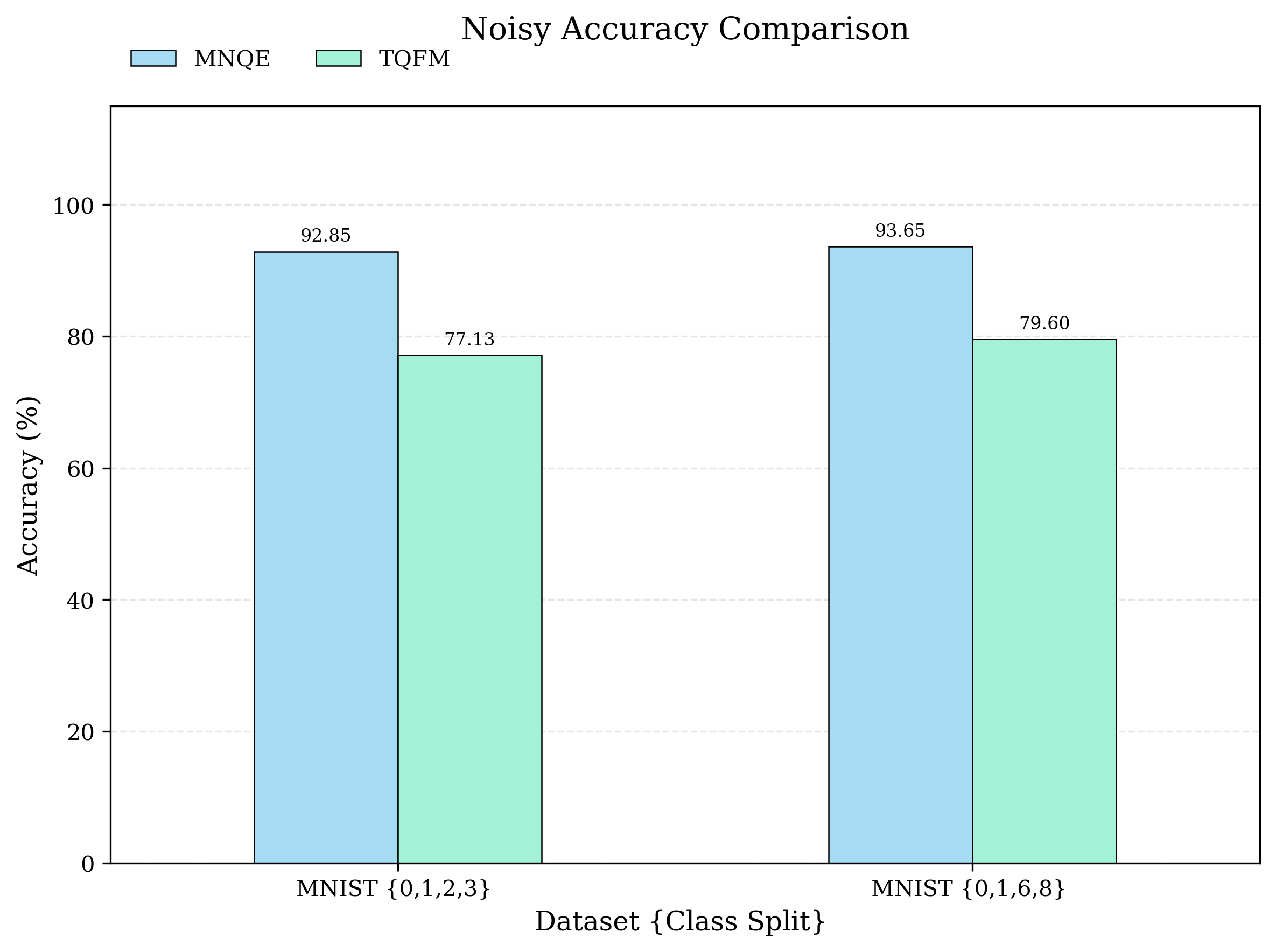}
        \caption{}
        \label{fig:16b}
    \end{subfigure}
    \caption{\justifying Comparison of the classification performance of different quantum embedding methods under noiseless and Gaussian-noise conditions. (a) Average classification accuracies of Without-MNQE, TQFM, MNQE, and PCA-MNQE on the MNIST $\{0123\}$, MNIST $\{0168\}$, Fashion-MNIST $\{0123\}$, and Fashion-MNIST $\{0168\}$ datasets under noiseless conditions. (b) Average classification accuracies of MNQE and TQFM on the MNIST $\{0123\}$ and MNIST $\{0168\}$ datasets under Gaussian noise perturbations. The values above the bars indicate the corresponding average classification accuracy (\%) of each model on the respective dataset.}
    \label{fig:1516}
\end{figure*}

Table~\ref{tab:results} summarizes the performance of the different methods under noiseless conditions in terms of accuracy, F1 score, recall and precision. Among the four dataset tasks, PCA-MNQE achieves the best overall performance, with all evaluation metrics exceeding those of the other methods, while MNQE achieves the second-best performance. To provide a more direct comparison of the classification accuracies, Fig.~\ref{fig:1516}\subref{fig:15a} presents the classification accuracies of the different models under noiseless conditions in the form of a bar chart, while Fig.~\ref{fig:1516}\subref{fig:16b} compares the classification accuracies of MNQE and TQFM under noisy conditions. The results show that models using MNQE maintain relatively high classification accuracy under noisy conditions, demonstrating good robustness to noise.
\begin{table}[!t]
\centering
\caption{Classification performance of different quantum models.}
\label{tab:results}
\begin{adjustbox}{max width=\columnwidth}
\begin{tabular}{lllcccc}
\toprule
Dataset & Class Split & Models & Accuracy (\%) & F1 Score (\%) & Recall (\%) & Precision (\%) \\
\midrule

\multirow[t]{8}{*}{MNIST}
& 0,1,2,3 & MNQE         & 94.10 & 94.08 & 94.10 & 94.09 \\
&         & \textbf{PCA-MNQE} & \textbf{96.25} & \textbf{96.28} & \textbf{96.25} & \textbf{96.50} \\
&         & Without MNQE & 31.65 & 31.11 & 31.65 & 31.02 \\
&         & TQFM        & 81.10 & 80.92 & 81.10 & 82.39 \\

& 0,1,6,8 & MNQE         & 94.15 & 94.18 & 94.15 & 94.29 \\
&         & \textbf{PCA-MNQE} & \textbf{96.35} & \textbf{96.45} & \textbf{96.45} & \textbf{96.47} \\
&         & Without MNQE & 32.85 & 31.59 & 32.85 & 31.75 \\
&         & TQFM        & 80.30 & 80.16 & 80.30 & 80.42 \\

\multirow[t]{8}{*}{Fashion-MNIST}
& 0,1,2,3 & MNQE         & 89.25 & 89.26 & 89.25 & 89.58 \\
&         & \textbf{PCA-MNQE} & \textbf{89.60} & \textbf{89.60} & \textbf{89.60} & \textbf{90.49} \\
&         & Without MNQE & 27.10 & 27.07 & 27.10 & 27.09 \\
&         & TQFM        & 79.03 & 79.08 & 79.05 & 79.04 \\

& 0,1,5,9 & MNQE         & 95.15 & 95.17 & 95.15 & 95.22 \\
&         & \textbf{PCA-MNQE} & \textbf{95.30} & \textbf{95.32} & \textbf{95.30} & \textbf{95.49} \\
&         & Without MNQE & 35.10 & 35.00 & 35.10 & 35.12 \\
&         & TQFM        & 83.05 & 82.74 & 83.05 & 83.76 \\

\bottomrule
\end{tabular}
\end{adjustbox}
\end{table}

\section{GENERALIZATION, EXPRESSIVITY AND TRAINABILITY OF MNQE}
\label{sec:5}

To further investigate the effects of the proposed MNQE method on the performance of quantum neural networks, this section examines three aspects: generalization, expressivity and trainability. By comparing the models with and without MNQE, we analyze its effects on the effective utilization of model parameters, the expressivity of quantum embeddings and the training process.

\subsection{Generalization Analysis Based on Local Effective Dimension}

Based on the preceding experimental results, the proposed MNQE method exhibits consistent performance improvements across different datasets and under both noiseless and noisy conditions. To further illustrate the MNQE from the perspective of model architecture, we introduce the local effective dimension (LED) proposed in Ref. \cite{A. Abbas1} to characterize the effective utilization of model parameters. This metric measures the effective number of parameter degrees of freedom that have a measurable influence on the model output for a given dataset, thereby providing a measure of the effective model complexity.

In the experiments, we use a four-qubit quantum neural network (QNN) as the reference model. The cases with and without MNQE are compared with fixed network architecture and training configuration. The experiments are performed independently for multiple training-set sizes using synthetically generated datasets. For each training-set size, the LED is calculated directly from the trained model.

\begin{figure}[!t]
    \centering
    \includegraphics[width=0.49\textwidth]{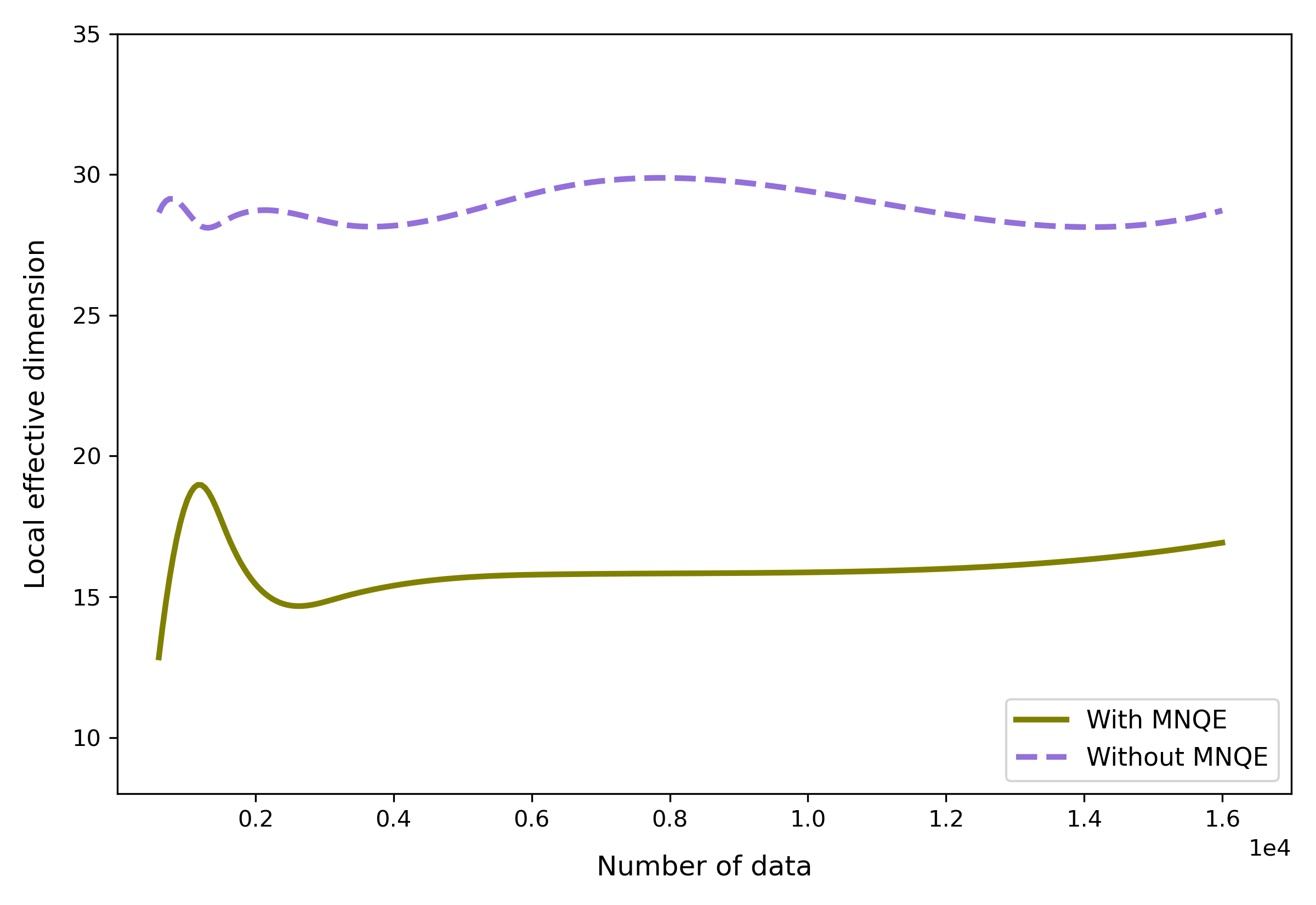}
    \caption{\justifying The figure compares the local effective dimension of the circuits with and without MNQE, represented by the green solid and blue dashed curves, respectively. All results are obtained from experiments on eight independently generated synthetic datasets.
}
    \label{fig:led}
\end{figure}

Fig.~\ref{fig:led} shows the variation of the LED with different training-set sizes. The green solid and purple dashed curves represent the local effective dimensions of the models with and without MNQE, respectively. The model incorporating MNQE consistently exhibits a lower LED across all training-set sizes. As the training-set size increases, the LED of both models increases, while the MNQE-based model maintains a more compact effective parameter structure. This behavior can be understood from the positive correlation between the LED and the generalization error: a smaller LED corresponds to a lower generalization error, and vice versa \cite{T. Hur}. These results indicate that, while retaining sufficient expressivity, MNQE imposes an effective structural constraint on the parameter optimization process, confining model training to a lower-dimensional effective subspace. This reduces the influence of redundant parameters and improves parameter utilization. Moreover, the lower effective dimension reduces the randomness of the parameter space, which may help alleviate barren plateaus and further improve the trainability and convergence stability of the model.

\subsection{Analysis of Expressivity and Trainability}

There is a trade-off between the expressivity and trainability in quantum neural networks. Higher expressivity enables a model to exhibit more complex quantum-state distributions, but may also increase the difficulty of optimization and reduce training stability \cite{J. R. McClean, Z. Holmes}. From an optimization perspective, higher expressivity is generally associated with more complex parameterized unitary structures, which can hinder the propagation of parameter perturbations to the measurement outputs, resulting in vanishing gradients and consequently slow or stalled convergence. This trade-off is particularly pronounced in learning models based on parameterized quantum circuits and directly affects their convergence behavior and final performance. Therefore, an appropriate balance between expressivity and trainability is essential in the design of quantum neural networks.

MNQE effectively alleviates the training difficulties associated with high expressivity. Specifically, MNQE learns a data-driven quantum embedding that structurally reshapes the distribution of quantum states, thereby enhancing the separability between classes. The key idea is to exploit prior information: a highly separable quantum embedding enhances the model's ability to approximate the underlying target mapping function. Without increasing the complexity of the quantum circuit, MNQE guides the quantum-state distribution toward a more discriminative structure, thereby alleviating the difficulty of training.

To quantitatively characterize the expressivity of the quantum embedding, we employ a measure based on the deviation from a unitary 2-design \cite{Z. Holmes, S. Sim}. Specifically, we consider the difference in the second-moment structure between the quantum-state ensemble and the Haar-random distribution,
\begin{equation}
A =
\int_{\mathrm{Haar}}
(|\psi\rangle\langle\psi|)^{\otimes 2} d\psi
-
\int_{E}
(|\phi\rangle\langle\phi|)^{\otimes 2} d\phi,
\end{equation}
where the first term represents the ideal second-order statistical structure under the Haar-random measure, while the second term represents the empirical distribution of the quantum-state ensemble $E$ generated by the data embedding. Based on this difference, the expressivity measure is defined as
\begin{equation}
\epsilon = \sqrt{\mathrm{Tr}(A^\dagger A)},
\end{equation}
where $\epsilon$ quantifies the deviation between the embedding distribution and the ideal 2-design. A smaller $\epsilon$ indicates that the quantum embedding is closer to the ideal random structure and therefore has greater expressivity, whereas a larger $\epsilon$ indicates more limited expressivity but greater structural distinguishability.

\begin{figure}[!t]
    \centering
    \includegraphics[width=0.45\textwidth]{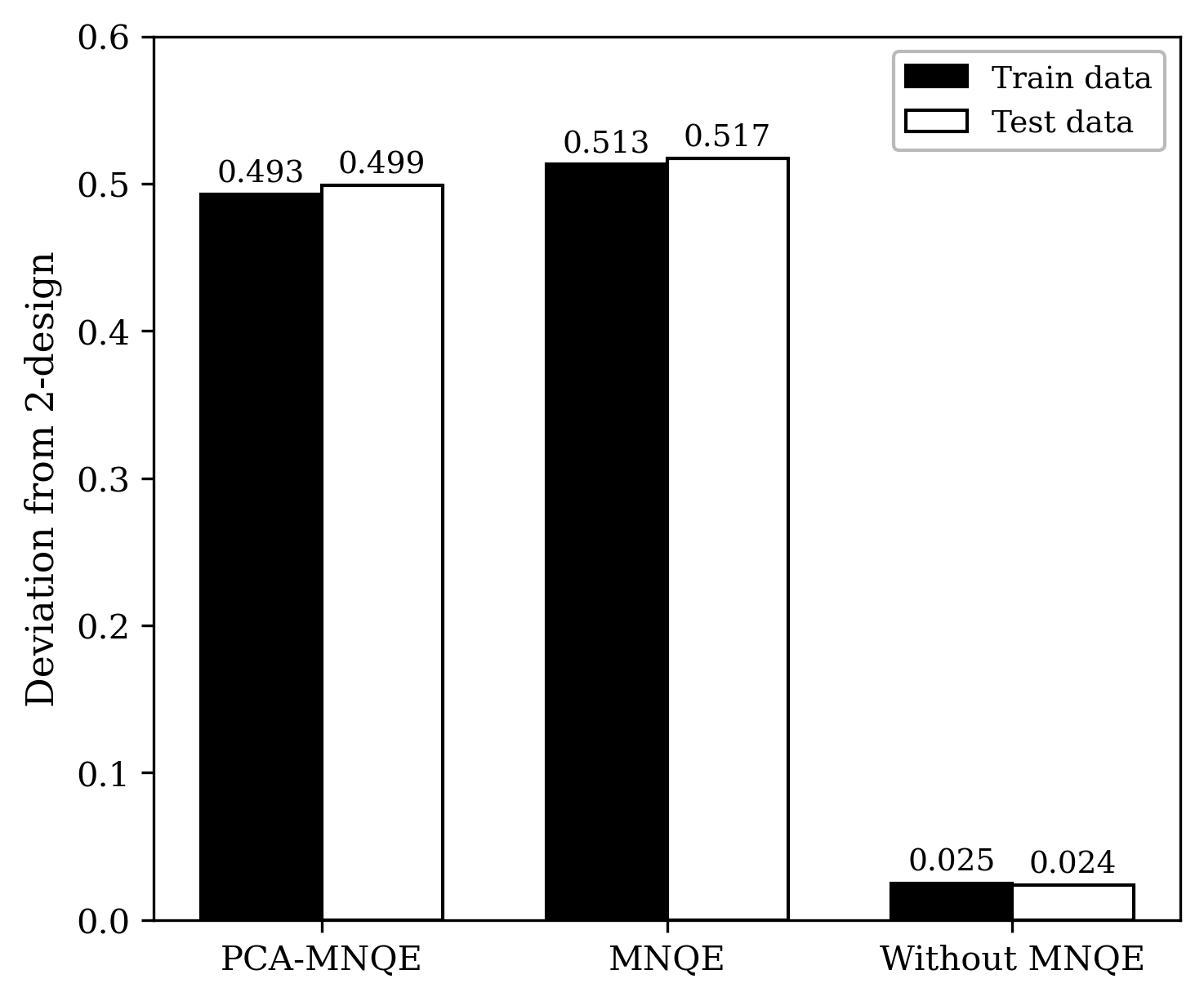}
    \caption{\justifying The expressivity of the models with and without MNQE is compared. The figure shows the deviation from a unitary 2-design, where a smaller deviation indicates greater expressivity. The deviation is calculated from the experimental results on the MNIST $\{0123\}$ training set (10,016 samples) and test set (2,000 samples).
}
    \label{fig:expressivity}
\end{figure}

The experimental results in Fig.~\ref{fig:expressivity} are obtained using the MNIST $\{0123\}$ dataset, with 10,016 training samples and 2,000 test samples. Three models are compared: MNQE, PCA-MNQE and Without-MNQE. The results show that introducing MNQE and PCA-MNQE increases the expressivity measure $\epsilon$, indicating that the expressivity of the quantum embedding is constrained to some extent. The models retain sufficient expressivity while exhibiting improved trainability.

\section{SUMMARY AND OUTLOOK}
\label{sec:6}

We have investigated the effect of the separability of quantum states from different classes on the performance of quantum classification models from the perspective of quantum-state discrimination, and derived a lower bound on the empirical risk for $K$-class quantum classification, establishing a theoretical connection between quantum-state separability and classification performance. Based on this analysis, we proposed a multiclass neural quantum embedding (MNQE) method for multiclass classification. Our MNQE introduces a classical neural network before quantum embedding to optimize the data representation, thereby enhancing the separability between quantum states from different classes after embedding. This reduces the lower bound on the empirical risk and improves classification performance. The effectiveness of MNQE is evaluated on multiclass classification tasks using the MNIST and Fashion-MNIST datasets. Comparing the trace distances between quantum states with and without MNQE shows that MNQE effectively increases the distances between quantum states from different classes, thereby improving their distinguishability. In terms of classification performance, both MNQE and PCA-MNQE outperform the conventional ZZ feature mapping and trainable quantum feature mapping methods, with PCA-MNQE achieving the best overall performance. Under noisy conditions, MNQE maintains high classification accuracy with relatively small performance degradation compared with the noiseless setting, demonstrating its robustness to noise. Furthermore, analyses of expressivity and local effective dimension show that MNQE reduces the effective model complexity and alleviates the optimization difficulties associated with excessive expressivity, achieving a more favorable balance between expressivity and trainability.

Our results may highlight future investigations on the effects of different neural network architectures and training strategies on MNQE performance. The integration of MNQE with structure learning \cite{M. Incudini} and adaptive quantum circuit design \cite{F. Zhang} may also be explored to further enhance the applicability of quantum machine-learning models to complex tasks and large scale quantum systems.

\begin{acknowledgments}
This work is being supported by the following: the National Natural Science Foundation of China (NSFC) under Grants 11761073, 12075159, and 12171044; the Academician Innovation Platform of Hainan Province.
\end{acknowledgments}

\appendix

\section{Derivation of Eq.~(9): Empirical Risk Lower Bound in Quantum Classification}
\label{app:risk_lower_bound}

This Appendix provides a detailed derivation of the empirical risk lower bound given in \eqref{eq:risk_lower_bound} for $K$-class quantum classification tasks.

Suppose that there are $K$ classes, with prior probability $p_i$ and the corresponding quantum state $\rho_i$ for class $i$. Given a set of POVM measurement operators $\{E_i\}_{i=1}^{K}$ satisfying $E_i \ge 0$ and $\sum_{i=1}^{K} E_i = I$, the success probability is given by
\begin{equation}
P_{\mathrm{succ}}=\sum_{i=1}^{K}p_i\,\mathrm{Tr}(E_i\rho_i),
\end{equation}
and the corresponding risk is $L_s = 1 - P_{\mathrm{succ}}$.

Define $\Delta_{ij}=p_i\rho_i-p_j\rho_j$ for $i\neq j$. Since $0\le E_i,E_j\le I$, it follows that $-I\le E_i-E_j\le I$. Using the variational characterization of the trace distance (see Appendix~\ref{app:trace_distance}), we obtain
\begin{equation}
\mathrm{Tr}\!\left[(E_i-E_j)\Delta_{ij}\right]
\le 2D_{\mathrm{tr}}(p_i\rho_i,p_j\rho_j).
\label{eq:trance}
\end{equation}
Let $D_{ij}=D_{\mathrm{tr}}(p_i\rho_i,p_j\rho_j)$. Eq.~\eqref{eq:trance} gives
\begin{equation}
p_i\mathrm{Tr}(E_i\rho_i)+p_j\mathrm{Tr}(E_j\rho_j)
- p_i\mathrm{Tr}(E_j\rho_i)-p_j\mathrm{Tr}(E_i\rho_j)
\le 2D_{ij}.
\label{eq:trance1}
\end{equation}
Rearranging Eq.~\eqref{eq:trance1}, we obtain
\begin{equation}
p_i\mathrm{Tr}(E_i\rho_i)+p_j\mathrm{Tr}(E_j\rho_j)
\le 2D_{ij}+p_i\mathrm{Tr}(E_j\rho_i)+p_j\mathrm{Tr}(E_i\rho_j).
\end{equation}
Summing over all pairs of classes satisfying $i<j$, the left-hand side becomes $(K-1)P_{\mathrm{succ}}$, while the distance terms sum to $2\sum_{i<j}D_{ij}$. Thus,
\begin{equation}
(K-1)P_{\mathrm{succ}} \le 2\sum_{i<j}D_{ij} + \sum_{i<j}\left[p_i\mathrm{Tr}(E_j\rho_i)+p_j\mathrm{Tr}(E_i\rho_j)\right].
\end{equation}
Using $\sum_{j\neq i}\mathrm{Tr}(E_j\rho_i)=1-\mathrm{Tr}(E_i\rho_i)$, we obtain
\begin{equation}
\begin{aligned}
\sum_{i<j}\!\left[p_i\mathrm{Tr}(E_j\rho_i)
+p_j\mathrm{Tr}(E_i\rho_j)\right]
&= \sum_{i=1}^{K}p_i\sum_{j\neq i}\mathrm{Tr}(E_j\rho_i) \\
&= \sum_{i=1}^{K}p_i\!\left(1-\mathrm{Tr}(E_i\rho_i)\right) \\
&= 1-P_{\mathrm{succ}}.
\end{aligned}
\end{equation}
It then follows that
\begin{equation}
(K-1)P_{\mathrm{succ}} \le 2\sum_{i<j}D_{ij} + (1 - P_{\mathrm{succ}}).
\end{equation}
Thus,
\begin{equation}
P_{\mathrm{succ}} \le \frac{1 + 2\sum_{i<j}D_{\mathrm{tr}}(p_i\rho_i,p_j\rho_j)}{K}.
\end{equation}
The resulting lower bound on the risk is given by
\begin{equation}
L_s \ge 1 - \frac{1 + 2\sum_{i<j}D_{\mathrm{tr}}(p_i\rho_i,p_j\rho_j)}{K}.
\end{equation}

\section{Derivation of the Variational Characterization of the Trace Distance}
\label{app:trace_distance}

This Appendix provides a derivation of the variational characterization of the trace distance used in Appendix~\ref{app:risk_lower_bound} and establishes its representation from the Hermitian variational characterization of the trace norm.

Let $A=A^\dagger$ be a Hermitian operator on a finite-dimensional Hilbert space. By the duality relation between the Schatten $1$-norm and the operator norm, we have \cite{J. Watrous}
\begin{equation}
\|A\|_1
=
\max_{\|B\|_\infty\le1}
\left|
\operatorname{Tr}(B^\dagger A)
\right|.
\label{eq:A1_dual}
\end{equation}

We next show that the original optimization problem and its restriction to the Hermitian feasible set have the same optimal value. Let $B$ be any feasible operator satisfying $\|B\|_\infty\le1$, and define
$z=\operatorname{Tr}(B^\dagger A)$.
If $z=0$, we simply take $H=0$. Assume $z\neq0$ and define
$C=e^{-i\arg(z)}B$.
Since $|e^{-i\arg(z)}|=1$, we have
$\|C\|_\infty=\|B\|_\infty\le1$,
and
\begin{equation}
\operatorname{Tr}(C^\dagger A)
=
e^{i\arg(z)}\operatorname{Tr}(B^\dagger A)
=
|z|.
\label{eq:phase_rotate}
\end{equation}
Decompose $C$ into its Hermitian and anti-Hermitian parts:
\begin{equation}
C=X+iY,~~
X=\frac{C+C^\dagger}{2},~~
Y=\frac{C-C^\dagger}{2i},
\end{equation}
where $X=X^\dagger$ and $Y=Y^\dagger$. Since $C^\dagger=X-iY$, we have
$\operatorname{Tr}(C^\dagger A)=\operatorname{Tr}(XA)-i\operatorname{Tr}(YA)$.
Because $A$, $X$ and $Y$ are all Hermitian operators,
$\operatorname{Tr}(XA),\operatorname{Tr}(YA)\in\mathbb{R}$.
Together with Eq.~\eqref{eq:phase_rotate}, this implies that the left-hand side is real, and hence
$\operatorname{Tr}(YA)=0$.
Therefore,
$|z|=\operatorname{Tr}(XA)$.
Using the convexity of the operator norm, we further obtain
\begin{equation} 
\|X\|_\infty = \left\| \frac{C+C^\dagger}{2} \right\|_\infty \le \frac{\|C\|_\infty+\|C^\dagger\|_\infty}{2} = \|C\|_\infty \le1. 
\end{equation}
Thus, $X$ is a Hermitian feasible operator satisfying $\|X\|_\infty\le1$, and for $A=A^\dagger$,
\begin{equation}
\operatorname{Tr}(XA)
=
\left|
\operatorname{Tr}(B^\dagger A)
\right|.
\end{equation}

Hence, for any feasible operator $B$, one can construct a Hermitian feasible operator $X$ with the same objective value, and then
\begin{equation}
\left|
\operatorname{Tr}(B^\dagger A)
\right|
=
\operatorname{Tr}(XA)
\le
\max_{\substack{H=H^\dagger\\\|H\|_\infty\le1}}
\operatorname{Tr}(HA).
\end{equation}
Since the above inequality holds for any feasible operator $B$ satisfying $\|B\|_\infty\le1$, taking the maximum of the left-hand side yields
\begin{equation}
\max_{\|B\|_\infty\le1}
\left|
\operatorname{Tr}(B^\dagger A)
\right|
\le
\max_{\substack{H=H^\dagger\\\|H\|_\infty\le1}}
\operatorname{Tr}(HA).
\end{equation}
On the other hand, since the set of Hermitian feasible operators is a subset of the original feasible set, we have
\begin{equation}
\max_{\substack{H=H^\dagger\\\|H\|_\infty\le1}}
\operatorname{Tr}(HA)
\le
\max_{\|B\|_\infty\le1}
\left|
\operatorname{Tr}(B^\dagger A)
\right|.
\end{equation}
Hence,
\begin{equation}
\|A\|_1
=
\max_{\substack{H=H^\dagger\\\|H\|_\infty\le1}}
\operatorname{Tr}(HA).
\label{eq:hermitian_variational}
\end{equation}
Since $H=H^\dagger$, the condition $\|H\|_\infty\le1$ is equivalent to $-I\le H\le I$. Therefore,
\begin{equation}
\|A\|_1
=
\max_{-I\le H\le I}
\operatorname{Tr}(HA).
\label{eq:trace_norm_variational}
\end{equation}

In particular, when $A=\rho-\sigma$, by the definition of the trace distance,
\begin{equation}
D_{\mathrm{tr}}(\rho,\sigma)
=
\frac12\|\rho-\sigma\|_1.
\end{equation}
we immediately obtain
\begin{equation}
D_{\mathrm{tr}}(\rho,\sigma)
=
\frac12
\max_{-I\le H\le I}
\operatorname{Tr}\!\left[
H(\rho-\sigma)
\right].
\label{eq:trace_distance_variational}
\end{equation}
Thus, for any Hermitian operator $H$ satisfying $-I\le H\le I$, we have
\begin{equation}
\operatorname{Tr}[H(\rho-\sigma)]
\le
2D_{\mathrm{tr}}(\rho,\sigma).
\end{equation}

\section{Relationship Between the Implicit Loss Function in Eq.~(12) and the Trace Distance}
\label{app:loss_trace_distance}

During the NQE training, we aim to enhance the separability between quantum-state ensembles associated with different classes. For a multiclass task, the quantum-state ensemble corresponding to class $c$ is defined as $\rho_c=\frac{1}{N_c}\sum_{i:y_i=c}|x_i\rangle\langle x_i|$, where $N_c$ denotes the number of samples in class $c$. Ideally, the class separability could be enhanced directly by increasing the trace distance between quantum-state ensembles of different classes. However, evaluating the trace distance can be computationally costly. Therefore, we adopt the fidelity-based implicit loss function in Eq.~\eqref{eq:loss} as the optimization objective.

When $y_i=y_j$, the loss function increases the fidelity $|\langle x_i|x_j\rangle|^2$ between samples from the same class. As shown in Eq.~\eqref{eq:same}, increasing the fidelity between samples within the same class increases the purity of the corresponding quantum-state ensemble,
\begin{equation}
\mathrm{Tr}(\rho_c^2)
=
\frac{1}{N_c^2}
\sum_{i,j:y_i=y_j=c}
|\langle x_i|x_j\rangle|^2,
\label{eq:same}
\end{equation}
thereby making the quantum states within the same class more concentrated in the Hilbert space.

When $y_i\neq y_j$, the loss function decreases the fidelity between samples from different classes. For any two classes $c$ and $d$, we assume $N_c=N_d=N$ for simplicity. By the strong convexity of the trace distance \cite{M. M. Wilde}, we have
\begin{equation}
D(\rho_c,\rho_d)
\le
\frac{1}{N}
\sum_{i=1}^{N}
\sqrt{
1-
|\langle x_i^{c}|x_i^{d}\rangle|^2
}.
\label{eq:diff}
\end{equation}
According to \eqref{eq:diff}, reducing the fidelity between samples from different classes increases the upper bound on the right-hand side, thereby promoting an increase in the trace distance between the corresponding quantum-state ensembles.

In summary, Eq.~\eqref{eq:same} shows that increasing the intraclass fidelity enhances the concentration of the corresponding class quantum-state ensemble, whereas inequality \eqref{eq:diff} indicates that reducing the interclass fidelity helps increase the trace distance between quantum-state ensembles of different classes. Therefore, although the trace distance is not directly optimized during training, the implicit loss function has an optimization direction consistent with increasing the trace distance between class quantum-state ensembles by simultaneously increasing the intraclass fidelity and decreasing the interclass fidelity. This demonstrates that fidelity can serve as an effective surrogate objective for the trace distance, enabling the optimization of quantum embeddings at a lower computational cost.

\section{Experimental Details}
\label{app:Details}

To enhance the separability of data from different classes in the quantum feature space, we perform comparative experiments using two neural quantum embedding methods, PCA-MNQE and MNQE. The experiments are conducted on the MNIST and Fashion-MNIST datasets. For the four-class classification tasks, a balanced subset is constructed by randomly sampling an equal number of samples from each of the four classes in each dataset. The training and test sets contain 10,016 and 2,000 samples, respectively.

For PCA-MNQE, principal component analysis (PCA) is first applied to the original data to extract $n$ principal components, where $n$ is the number of qubits used in the quantum embedding circuit. The reduced features are then fed into a fully connected neural network with two hidden layers. In the eight-qubit setting, each hidden layer contains 16 neurons, while the output layer has dimension $n$, with its output used to parameterize the quantum embedding circuit. The ReLU activation function is used throughout the network. For MNQE, a two-dimensional convolutional neural network is directly applied to the original images for feature extraction. A max-pooling layer follows each convolutional layer to progressively reduce the spatial dimensions of the feature maps. After two successive convolution and max-pooling operations, the spatial dimensions are reduced from \(28 \times 28\) to \(14 \times 14\) and then to \(7 \times 7\), before the resulting features are mapped to an \(n\)-dimensional feature vector. The classical neural networks in both embedding models are trained by minimizing the implicit fidelity loss $L$ defined in Eq.~\eqref{eq:loss}, thereby enhancing the separability of quantum states from different classes.

During optimization, stochastic gradient descent \cite{Y. A. LeCun} is used for the quantum embedding stage, with 300 training epochs, a learning rate of 0.005, and a batch size of 32. The quantum embedding circuit consists of three layers of the ZZ feature map. During the training of the ansatz layer, the quantum classifier is optimized using the Adam optimizer \cite{D. P. Kingma}, with 500 training epochs, a learning rate of 0.01, and a batch size of 128. The ansatz layer follows the circuit architecture shown in Fig.~\ref{fig:circuit2} and is repeated for three layers. All experimental results are obtained using the above unified parameter settings.

\end{document}